%% file: main.tex
\documentclass{article}

\usepackage{PRIMEarxiv}

\usepackage[utf8]{inputenc} 
\usepackage[T1]{fontenc}    
\usepackage{hyperref}       
\usepackage{url}            
\usepackage{booktabs}       
\usepackage{amsfonts}       
\usepackage{nicefrac}       
\usepackage{microtype}      
\usepackage{lipsum}
\usepackage{fancyhdr}       
\usepackage{graphicx}       
\graphicspath{{media/}}     
\usepackage{algorithm}
\usepackage{algpseudocode}
\usepackage{multirow}
\usepackage{amsmath}
\usepackage{amssymb}
\usepackage{mathtools}
\usepackage{amsthm}
\usepackage{subcaption}

\title{Rethinking Oversaturation in Classifier-Free Guidance via Low Frequency
}

\author{
Kaiyu Song, \\
  Sun Yat-Sen University \\ \\
   \And
  Hanjiang Lai \\
  Sun Yat-Sen University \\\\
}

\begin{document}
\maketitle

\input{sec/0_abs}
\input{sec/1_intro}

\input{sec/2_related_work}
\input{sec/3_pre}

\input{sec/4_method}
\input{sec/5_experiment}

\input{sec/6_conclusion}

\input{sec/x_supply}

\bibliographystyle{unsrt}  
\bibliography{main}

\end{document}

%% file: sec/0_abs.tex
\begin{abstract}
Classifier-free guidance (CFG) succeeds in condition diffusion models that use a guidance scale to balance the influence of conditional and unconditional terms. A high guidance scale is used to enhance the performance of the conditional term. However, the high guidance scale often results in oversaturation and unrealistic artifacts. In this paper, we introduce a new perspective based on low-frequency signals, identifying the accumulation of redundant information in these signals as the key factor behind oversaturation and unrealistic artifacts. Building on this insight, we propose low-frequency improved classifier-free guidance (LF-CFG) to mitigate these issues. Specifically, we introduce an adaptive threshold-based measurement to pinpoint the locations of redundant information. We determine a reasonable threshold by analyzing the change rate of low-frequency information between prior and current steps. We then apply a down-weight strategy to reduce the impact of redundant information in the low-frequency signals. Experimental results demonstrate that LF-CFG effectively alleviates oversaturation and unrealistic artifacts across various diffusion models, including Stable Diffusion-XL, Stable Diffusion 2.1, 3.0, 3.5, and SiT-XL.
\end{abstract}

%% file: sec/1_intro.tex
\section{Introduction}
\label{sec:intro}
Conditional generation-based diffusion models~\cite{sd} aim to generate samples aligned with a condition like text prompts. This is essential for various tasks, e.g., text-to-image~\cite{sd3}, text-to-video~\cite{opensora}, and more.

Recent works~\cite{class-based-cfg,cfg,baseline} have demonstrated the success of conditional generative methods in reusing well-trained unconditional diffusion models. Leveraging the potential of these models, classifier-free guidance (CFG)~\cite{cfg} has achieved state-of-the-art performance in various conditional generation tasks, where a conditional term is incorporated into the diffusion process. The unconditional term reuses pre-trained unconditional diffusion models to maintain the diversity of the generated samples. CFG uses a guidance scale to control the balance between conditional and unconditional terms, offering a trade-off between alignment and diversity.

\begin{figure}
    \centering
    \includegraphics[width=0.9\linewidth]{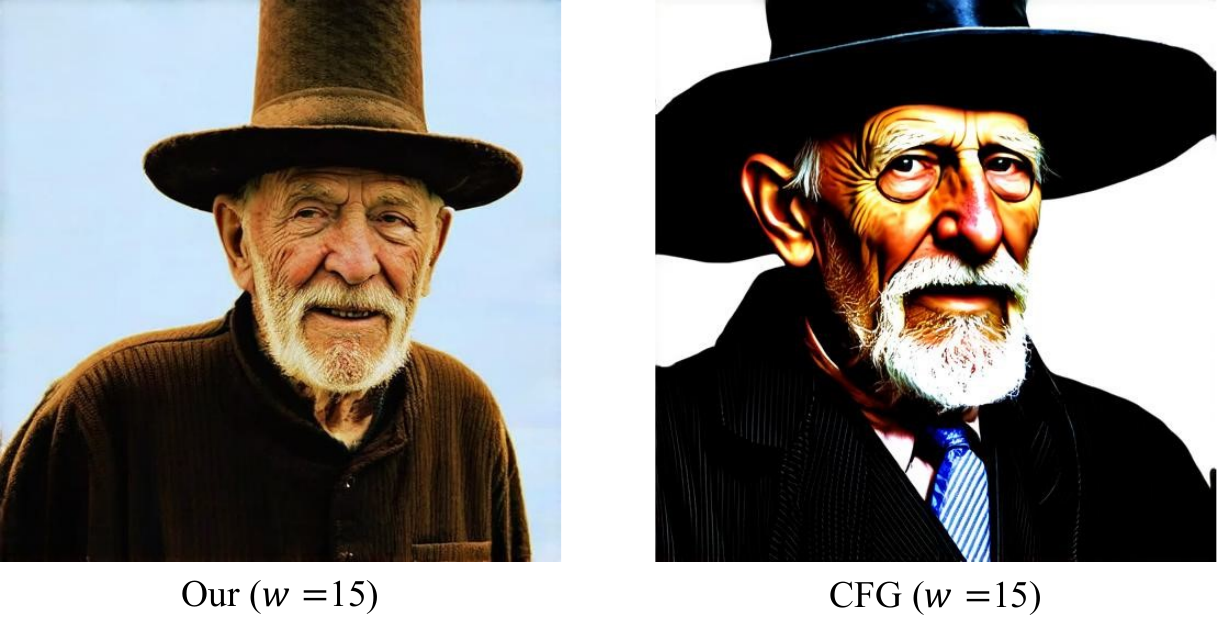}
    \caption{An illustration of oversaturation, where $w$ is the guidance scale. By comparing the generated image under a high guidance scale between our method and CFG, it can be found that some pixel colors (e.g., the face part) in the images from CFG will be over-exposed (e.g., the image brightness is too bright).}
    \label{fig:oversaturation}
\end{figure}

Despite its effectiveness, CFG suffers from an oversaturation problem when the guidance scale is high~\cite{baseline}. To illustrate this issue, we show the text-to-image task in Fig.~\ref{fig:oversaturation}. Images generated with a high guidance scale can appear abnormal, with pixel colors tending towards over-exposure. 

Previous methods~\cite{selfguide,pag,cfgpp} tend to avoid oversaturation by improving the guidance scale and limiting the range of the guidance scale. In contrast to the strategy of avoidance and careful tuning, recent methods such as APG~\cite{baseline} were proposed. APG re-formulates the conditional and unconditional terms into a gradient descent representation, allowing it to be split into parallel and orthogonal directions. This decoupling strategy then alleviates oversaturation by down-weighting the parallel gradient direction. However, the underlying cause is still unknown.

In this paper, we offer a novel perspective based on low-frequency signals to provide an intuitive explanation for oversaturation. We observe that redundant information accumulates within low-frequency signals, leading to oversaturation. Building on this insight, we propose low-frequency improved classifier-free guidance (LF-CFG) to mitigate redundant information in low-frequency signals.

Specifically, we first extract low-frequency signals for both unconditional and conditional terms using a linear-based low-frequency filter. We then introduce an adaptive threshold-based measurement to identify the locations of redundant information. Following our insight, redundant information accumulates in signals with similar updated values~\cite{cfg}. We recognize such redundant information by analyzing the change rate of low-frequency information between prior and current steps. We then calculate a threshold adaptively based on the statistical rule of the change rate. A novel down-weighting strategy is then applied to alleviate redundant information in low-frequency signals. Experimental results demonstrate that LF-CFG effectively reduces redundant information and oversaturation, thereby improving the quality of generated images.

Our main contributions are summarized as follows:

\begin{itemize}
    \item We propose a novel LF-CFG, an improved CFG for removing the redundant information hidden in the low-frequency signal.
    \item We offer a novel view to explain the oversaturation problem in CFG based on the frequency theory.
    \item The experimental results on five popular diffusion models demonstrate that LF-CFG could achieve SOTA performance and improve the generation quality on a high guidance scale.
\end{itemize}

%% file: sec/2_related_work.tex
\section{Related Work}
\label{sec:related}
\textbf{Conditional generation based on diffusion models.} It introduces the condition, such as the text prompt, to control the generation process of diffusion models. For example, stable diffusion~\cite{sd,sd3} introduced the text prompt to guide image generation. OpenSora~\cite{video_gen} introduced the text prompt to guide video generation.

CFG~\cite{cfg}, as one of the most popular technologies to improve the generation quality, has been widely implemented in various conditional generation tasks such as text-image generation~\cite{sd,sd3}, text-to-video generation~\cite{video_gen}, and text-to-3D generation~\cite{text-to-3d}. CFG has been proposed based on the classifier-based guidance~\cite{class-based-cfg}, where the classifier has been replaced by directly introducing conditions into the backbone of the diffusion models.

\textbf{Improvement for classifier-free guidance.} Previous works~\cite{cads,baseline,ifg} found that a higher guidance scale causes the oversaturation phenomenon and thus decreases the Frechet inception distance (FID)~\cite{fid}. To alleviate this, the methods improve CFG in a limited range of guidance scale, thus avoiding leading to the oversaturation phenomenon. For example, CADS~\cite{cads} proposed to adjust the noise schedule to increase the diversity under the high guidance scale to improve the FID. Independent condition guidance (IFG)~\cite{ifg} proposed an additional time scheduler guidance to improve the diversity. PAG~\cite{pag} proposed to perturb the attention to improve the low-level CFG. Tero \textit{et al.}~\cite{selfguide} proposed the self-guided CFG by leveraging the bad version of models. Chung \textit{et al.}~\cite{cfgpp} proposed the CFG++ by introducing manifold theory.

Recently, Robin \textit{et al.}~\cite{baseline} first introduced gradient theory to analyze the phenomenon of oversaturation, which divided the CFG into orthogonal and parallel signals. They successfully alleviated the oversaturation by weakening parallel signals. This opens a way to explore the root cause of the oversaturation. Building upon the need for a deeper understanding, our work proposes LF-CFG. Instead of a gradient analysis, we investigate the oversaturation phenomenon through the frequency theory. LF-CFG demonstrates that the accumulation of redundant information within low-frequency signals may be the possible the underlying root cause. LF-CFG specifically targets this identified cause to mitigate oversaturation, offering a complementary perspective and solution.

%% file: sec/3_pre.tex
\section{Preliminary}
\begin{figure*}
    \centering
    \includegraphics[width=0.9\linewidth]{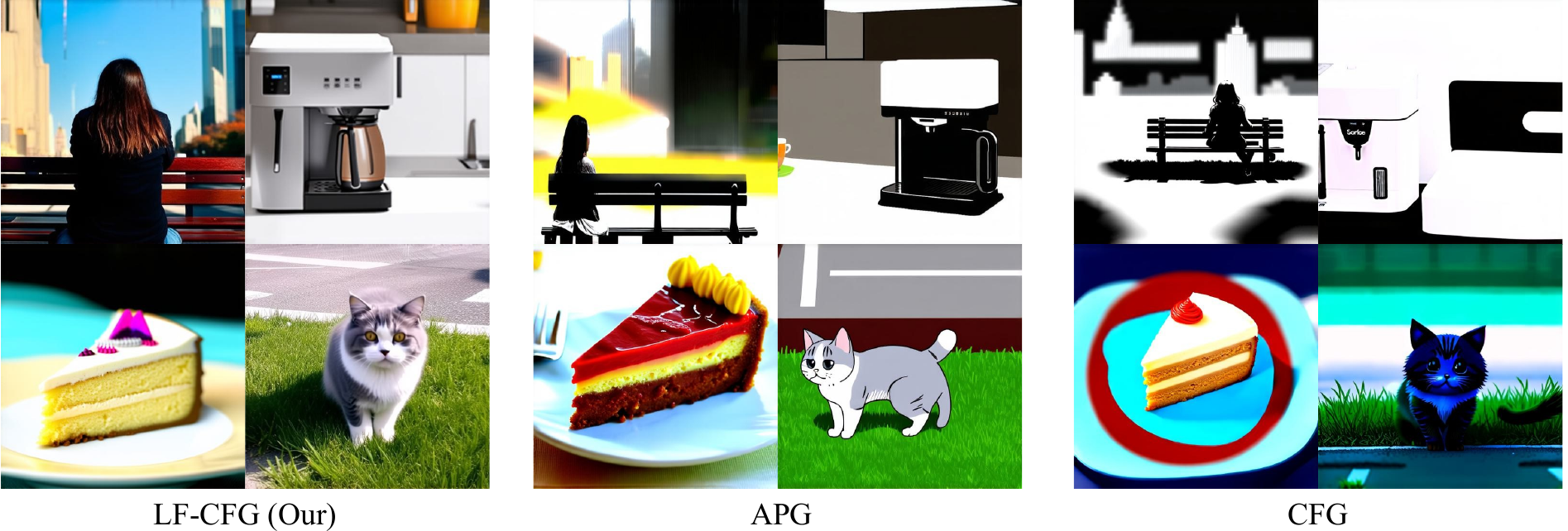}
    \caption{Qualitative results based on Stable Diffusion 3 ($w=15$)}
    \label{fig:sd3}
\end{figure*}
\label{sec:pre}
\textbf{CFG.} CFG aims at incorporating the condition into the reverse process of the diffusion models. Concretely, given a pre-trained unconditional diffusion model $v_{\theta}(x_{t},t)$,  where $\theta$ is the trained parameters of the diffusion models, $t$ is the $t$-th timestep within the time interval $[0, T]$ started from $T$ to the $0$. The reverse process is defined as follows:
\begin{equation}
    dx_{t} = v_{\theta}(x_{t},t)dt.
    \label{eq:reverse}
\end{equation}
Please note that Eq.~\ref{eq:reverse} is the flow-matching~\cite{flow-match} based formation, which can be converted to the SDE~\cite{sde} and DDIM~\cite{ddim}. Then, CFG can be re-formulated the Eq.~\ref{eq:reverse} as follows:
\begin{equation}
    dx_{t} = [v_{uc}(t) + w*(v_{c}(t) - v_{uc}(t))]dt,
    \label{eq:cfg}
\end{equation}
where $w$ is the guidance scale. $v_{c}(t)=v_{\theta}(x_{t},t,c)$ is the conditional diffusion model, and $v_{uc}(t) = v_{\theta}(x_{t},t,\emptyset)$ is unconditional term. To simplify the notation, we use $v_{c}(t)$ and $v_{uc}(t)$ to represent the conditional and unconditional terms at $t$-th timesteps, respectively. A challenge is that the generation quantity of the CFG will tend to be oversaturated when $w$ tends to be a high value.
\begin{figure*}[h!]
    \centering
    \includegraphics[width=0.9\linewidth]{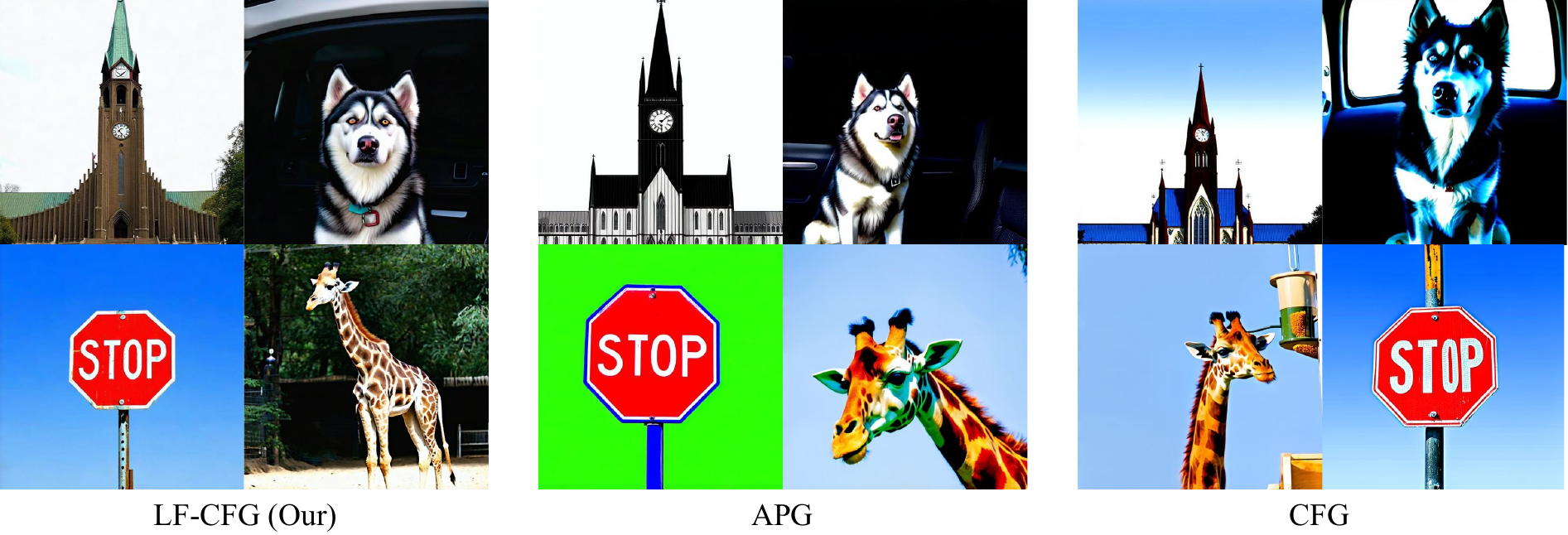}
    \caption{Qualitative results based on Stable Diffusion 3.5 ($w=15$)}
    \label{fig:sd3.5}
\end{figure*}

\textbf{Frequency decoupling.} Given a low-frequency filter $f_{l}$ and high-frequency filter $f_{h}$, we could get the high- and low-frequency signal for unconditional term and conditional term as follows:
\begin{equation}
    \begin{split}
        v^{l}_{c}(t) = f_{l}(v_{c}(t)), 
        v^{l}_{uc}(t) = f_{l}(v_{uc}(t)), 
        \\
        v^{h}_{c}(t) = f_{h}(v_{c}(t)), 
        v^{h}_{uc}(t) = f_{h}(v_{uc}(t)),
    \end{split}
    \label{eq:low_freq}
\end{equation}
where $ v_{c}^{l}(t)$ and $v_{uc}^{l}(t)$ are the low-frequency signal from $ v_{c}(t)$ and $v_{uc}(t)$, respectively. $v^{h}_{c}(t)$ and $v_{uc}^{h}(t)$ are the high-frequency signals from $ v_{c}(t)$ and $v_{uc}(t)$, respectively. $ v^{*}_{*}(t) \in \mathbb{R}^{ C \times H \times W}$ represents the dimension of the signals. $C$ is the channel. $H$ and $W$ are the resolutions.

%% file: sec/4_method.tex
\section{Methodology}
\label{sec:method}

\begin{figure}
    \centering
    \includegraphics[width=0.95\linewidth]{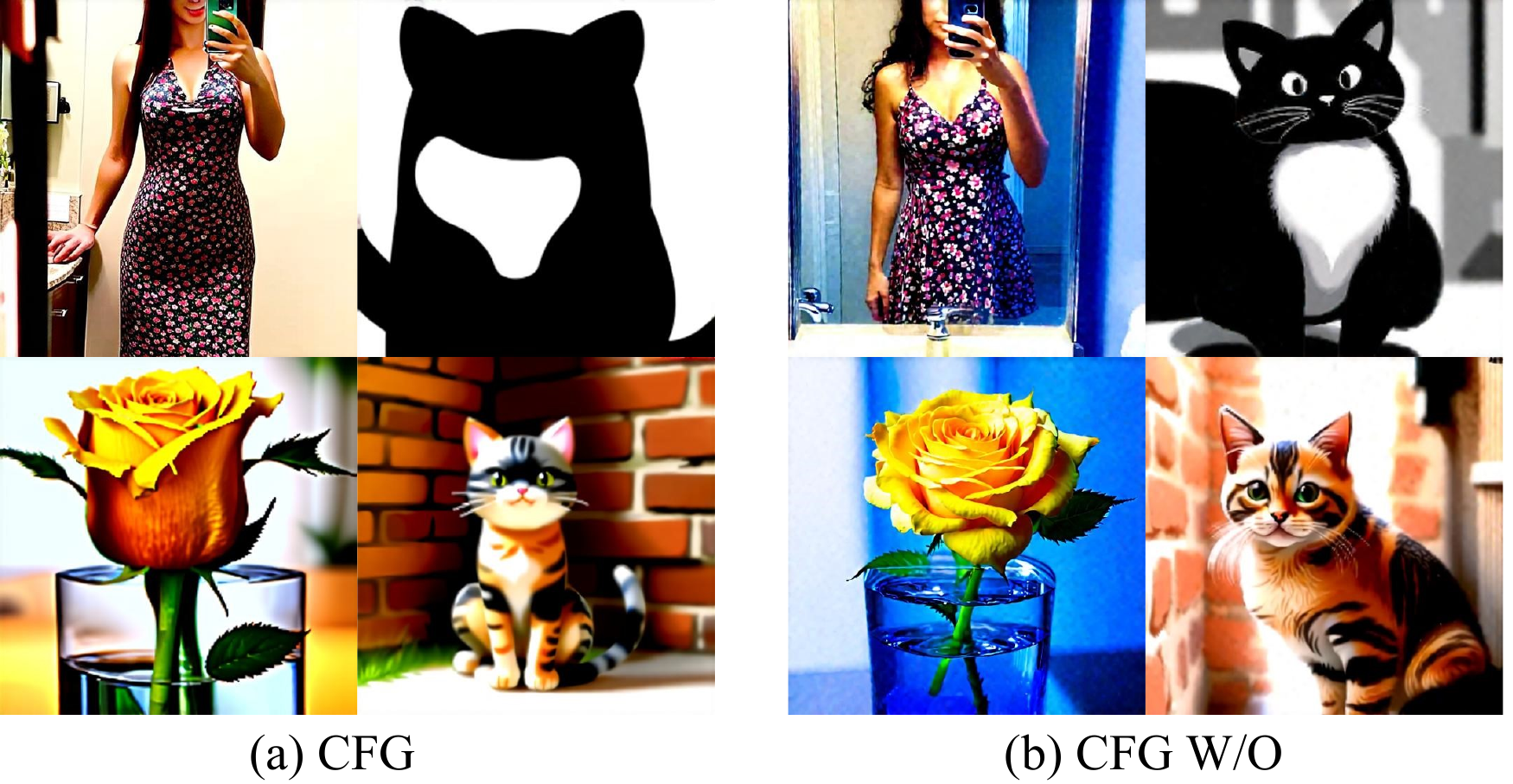}
    \caption{An illustration of high-frequency for the oversaturation generated by different prompts, where W/O represents no high-frequency signal. Without the high-frequency signal, generated images still suffer from oversaturation.}
    \label{fig:removing_high}
\end{figure}

\begin{figure}
    \centering
    \includegraphics[width=0.9\linewidth]{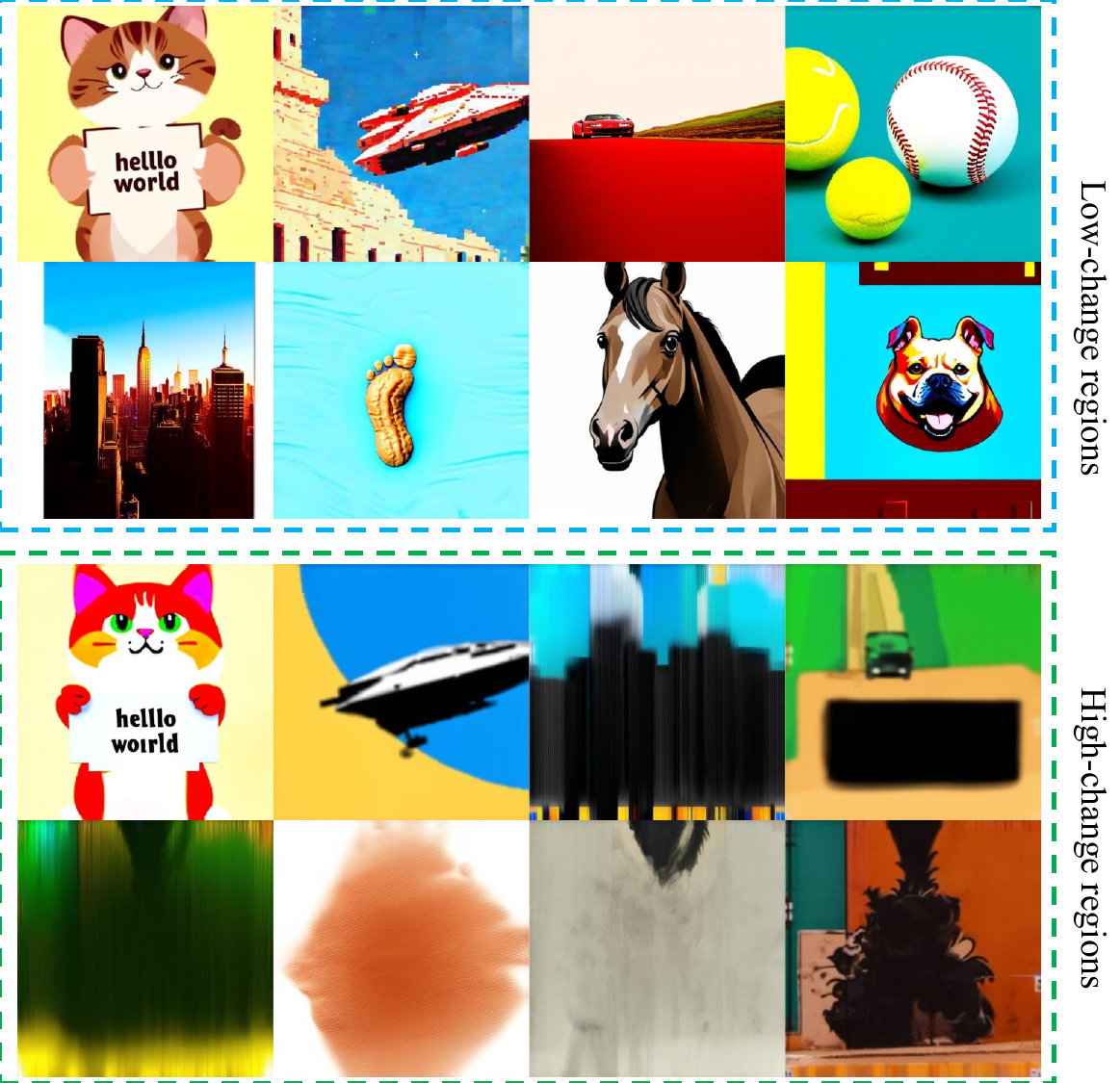}
    \caption{An illustration of a low-frequency signal for oversaturation, where the low-change region represents generating images by zeroing the low-change region. The high-change region represents generating images by zeroing the high-change regions. The results show that zeroing the low-region changes region could alleviate oversaturation while keeping the generation quality. }
    \label{fig:evidence}
\end{figure}

\subsection{What Causes the Oversaturation?}

\textbf{Low-frequency or high-frequency?} Our investigation into the oversaturation phenomenon in CFG begins with low- and high-frequency. In image generation and processing, high-frequency signals capture fine details and textures, often exhibiting significant changes during the reverse process. Conversely, low-frequency signals represent the global structure, semantics, and smoother color transitions~\cite{hipa}, which remain stable during the reverse process.

Oversaturation results in large regions of extreme, unrealistic color values, distorting global characteristics. Thus, low frequency signals tend to be the primary contributors to the oversaturation artifact observed at high guidance scale $w$.

To prove this, we conducted an experiment explicitly removing the high-frequency components during the reverse process. Specifically, we decompose the unconditional $v_{uc}$ and conditional $v_{c}$ at each step into low-frequency $v^l_{j}$ and high-frequency $v^h_{j}$ components using a linear filter, where $j \in \{uc,c\}$ We then set the high-frequency components to zero ($v^h_{uc} = 0, v^h_{c} = 0$) before applying the CFG. As shown in Fig~\ref{fig:removing_high} (right panels), the generated images still suffer from significant oversaturation, although some fine details are lost. This confirms that the low-frequency components predominantly lead to oversaturation. Mitigating it requires focusing on how these components ($v^l_{uc} = 0, v^l_{c} = 0$) are handled in CFG.

\textbf{Problematic regions in low-frequency.} We further investigate how the low-frequency signal leads to oversaturation. Our intuition is that regions where the low-frequency signal changes very little over time might accumulate excessive information, i.e., redundant information. A high guidance scale will accumulate this redundant information, leading to saturation. Thus, we propose analyzing the temporal dynamics by examining how much these signals change from one step of the reverse process to the next.
\begin{equation}
    \begin{split}
        r^{l}_{uc}(t) &= d(v^{l}_{uc}(t),v^{l}_{uc}(t+1)) \\
        r^{l}_{c}(t) &= d(v^{l}_{c}(t),v^{l}_{c}(t+1)),
    \end{split}
    \label{eq:rate}
\end{equation}
where $d(\cdot, \cdot)$ represents the pixel-wise Euclidean distance, calculated independently for each spatial location across all channels. The resulting $r^{l}_{j}(t)$ (where $j \in \{uc, c\}$) is a map indicating the magnitude of change at each location.

To quantify this, we measure the change rate for the low-frequency unconditional $v^l_{uc}$ and conditional $v^l_{c}$ signals between the current step $t$ and the previous step $t+1$ shown in Eq.~\ref{eq:rate}.

We then categorize the spatial locations based on the change rate relative to the overall statistics of the change map at step $t$. The change rates $r^{l}_{j}(t)$ across all locations approximately follow a distribution amenable to mean/std analysis (reasonable given our use of linear filters). Thus, we define two region types using a threshold based on one standard deviation:
\begin{equation}
\begin{split}
   r^{l}_{j}(t) &< \text{mean}(r^{l}_{j}(t))- \text{std}(r^{l}_{j}(t)), \text{Low-change region},\\
    r^{l}_{j}(t) &> \text{mean}(r^{l}_{j}(t)) + \text{std}(r^{l}_{j}(t)), \text{High-change region}.
     \label{eq:measurement}
\end{split}
\end{equation}
where, $\text{mean}(\cdot)$ and $\text{std}(\cdot)$ are calculated over all spatial locations for the map $r^{l}_{j}(t)$. The low-change regions correspond to areas where the low-frequency signal is relatively static between steps, potentially accumulating redundant information. The threshold $\text{mean} + \text{std}$ serves as a practical heuristic to identify the lower tail.

We also perform another set of zeroing experiments to verify this, focusing only on the low-frequency signals. 1) Zeroing low-change regions. We set the low-frequency signal values $v^{l}_{j}(t)$ to 0 at locations identified as low-change regions by Eq.~\ref{eq:measurement}. 2) Zeroing high-change regions. We set $v^{l}_{j}(t)$ to 0 at locations identified as high-change regions.

The results are reported in Fig.~\ref{fig:evidence}. Zeroing the low-change regions successfully alleviates oversaturation, resulting in visually plausible images. Conversely, zeroing the high-change regions fails to prevent oversaturation and can lead to generation failure. This proves that the accumulation of redundant information of low-frequency signals is a possible root cause of the oversaturation.

\textbf{Discussion.} Our motivation for such a category is clear: Finding which signal mainly causes the oversaturation problem. We first decouple the output of the diffusion models into the high- and low-frequency signals. We find that low-frequency signals are the primary contributors to oversaturation. Then, to explore the root cause further, we categorize the low-frequency signal into the low- and high-change regions according to the change rate between previous and current steps in diffusion models.

There is an example to explain why the low-change region is a problem. During the reverse diffusion process, a pixel's final value is determined by accumulating low-frequency signal adjustments across timesteps: $\Delta=\{r^{l}_{j}(1),...,r^{l}_{j}(t)\}$. We assume the initial pixel value is $0.5$. The final value is given by $\text{Final Value} = 0.5 + w\sum_{t=1}^{T}r^{l}_{j}(t)$, where $w$ is the CFG scale. If adjustments are small and consistent, their cumulative effect becomes dominant.

We define bounds for low-change regions $|r^{l}_{j}(t) - r^{l}_{j}(t+1)|_{2}< \text{mean} - \text{std}$ and high-change regions $|r^{l}_{j}(t) - r^{l}_{j}(t+1)|_{2}> \text{mean} + \text{std}$. Adjusting a low-change region that is slowly varying and additive can cause oversaturation. For example, if $r^{l}_{j}(1) = r^{l}_{j}(2) ... = r^{l}_{j}(T) \approx +0.1$. The cumulative sum becomes $0.5 + w*T*0.1$. At high guidance scales, this positive feedback loop forces pixel values toward extremes (e.g., over-exposed whites). Conversely, high-change regions exhibit self-correcting behavior due to larger, fluctuating adjustments (e.g., $r^{l}_{j}(1) = +0.3$, $r^{l}_{j}(2) = -0.2$), preventing systematic accumulation. 

\subsection{LF-CFG} 
Oversaturation arises from accumulating low-frequency signals in regions of low temporal change. Leveraging this, we propose low-frequency improved classifier-free guidance (LF-CFG). This method adaptively identifies these problematic regions and applies a targeted down-weighting strategy to mitigate the accumulation effect without discarding essential global information. LF-CFG consists of two main components: an adaptive measurement of low-change regions and a specific down-weighting strategy integrated into CFG.

\textbf{Adaptive measurement.} To precisely locate regions susceptible to redundant information accumulation, we first calculate the step-wise change rate maps $r^{l}_{uc}(t)$ and $r^{l}_{c}(t)$ for low-frequency components, as defined in Eq.~\ref{eq:rate}.

We then define a mask, $m^{l}_{j}(t)$, to identify the low-change regions spatial locations, where the change rate is below the average. Based on our design of linear frequency filters, the change rates $r^{l}_{j}(t)$ tend to be a Gaussian distribution. This property enables us to define the threshold $\gamma^{l}_{j}(t)$ based on the lower tail of the distribution $\gamma^{l}_{j}(t) = \text{mean}(r^{l}_{j}(t)) + \text{std}(r^{l}_{j}(t))$.
We also explore different tails of the distribution e.g., $\text{mean}(r^{l}_{j}(t)) - \text{std}(r^{l}_{j}(t))$ in the ablation study and find $\text{mean}(r^{l}_{j}(t)) + \text{std}(r^{l}_{j}(t))$ performs best. 

$\gamma^{l}_{j}(t)$ aims to capture approximately 66.7\% of the locations with the lowest change rate, assuming a quasi-Gaussian distribution. The binary mask $m^{l}_{j}(t) \in \{0,1\}^{C \times H \times W}$ for $j \in \{c, uc\}$ is then generated, where a value of 1 indicates a location within the low-change region:
\begin{equation}
    m^{l}_{j}(t)= \begin{cases} 1 & \text{if } r^{l}_{j}(t) < \gamma^{l}_{j}(t) \\ 0 & \text{otherwise} \end{cases}.
    \label{eq:mask_definition}
\end{equation}
Thus, $m^{l}_{j}(t)$ targets the spatial locations in the low-frequency signals considered likely to accumulate redundant information susceptible to oversaturation when amplified by high guidance.

\textbf{Down-weight strategy.} After obtaining $m^{l}_{j}(t)$, we propose a straightforward approach to mitigate redundant information. We can effectively weaken the low-frequency signals by calculating the ratio between the low- and high-change regions. Specifically, we introduce a weight $\rho$ to represent the proportion of the low- and high-change regions. We then re-weight the low-frequency signals as follows: 
\begin{equation}
    \begin{split}
        v^{l}_{j} = \rho v^{l}_{j} m^{l}_{j}(t) & + v^{l}_{j} (1-m^{l}_{j}(t))  \\
        s.t. j & \in \{c,uc\}.
    \end{split}
    \label{eq:new_freq}
\end{equation}
Given that approximately 66.7\% of the value of $v^{l}_{j}$ is in low-change regions and 33.3\% is in high-change regions, respectively. The proportion $\rho$ could be calculated $\rho = 33.3\%/66.7\%$. This is the reason for choosing the linear map to filter the low-frequency. The statistical principle could help us simplify the design of the hyperparameter. Thus, LF-CFG could work on various diffusion models.

\textbf{Integrating down-weighting into CFG.} We have the mechanism to create modified low-frequency signals $v^{l*}_{uc}(t)$ and $v^{l*}_{c}(t)$ that suppress potentially redundant information. There is a question of how these modified signals can be integrated into the CFG.

To illustrate this problem, we simplify the CFG based on the low-frequency signals from Eq.~\ref{eq:cfg} as follows:
\begin{equation}
    dx_{t} \approx [v^{l}_{uc} + w*(v^{l}_{c} - v^{l}_{uc})]dt. \label{eq:freq_flow_repeat_again}
\end{equation}
This structure involves the unconditional term $v^{l}_{uc}$ and the difference term $(v^{l}_{c} - v^{l}_{uc})$. We cannot simply replace all $v^{l}_{j}$ with $v^{l*}_{j}$, as different terms might play different roles. For example, modifying $v^{l}_{uc}$ lonely might affect the overall structure differently than modifying the terms within the difference term.

This leads to a combination problem about how best to substitute $v^{l}_{uc}$ and $v^{l}_{c}$ with their modified versions $v^{l*}_{uc}$ and $v^{l*}_{c}$ within Eq. \ref{eq:freq_flow_repeat_again}? We explored four primary combinations for applying the modification: 1) $v^{l*}_{uc} + w(v^{l*}_{c} - v^{l*}_{uc})$. Modify both terms. 2) $v^{l}_{uc} + w(v^{l*}_{c} - v^{l}_{uc})$. Modify $v^{l}_{c}$ within the difference term. 3) $v^{l}_{uc} + w(v^{l*}_{c} - v^{l*}_{uc})$. Modify both $v^{l}_{c}$ and $v^{l}_{uc}$ within the difference term. 4) $v^{l*}_{uc} + w(v^{l}_{c} - v^{l*}_{uc})$. Modify $v^{l}_{uc}$ in all terms.
\begin{figure}
    \centering
    \includegraphics[width=0.9\linewidth]{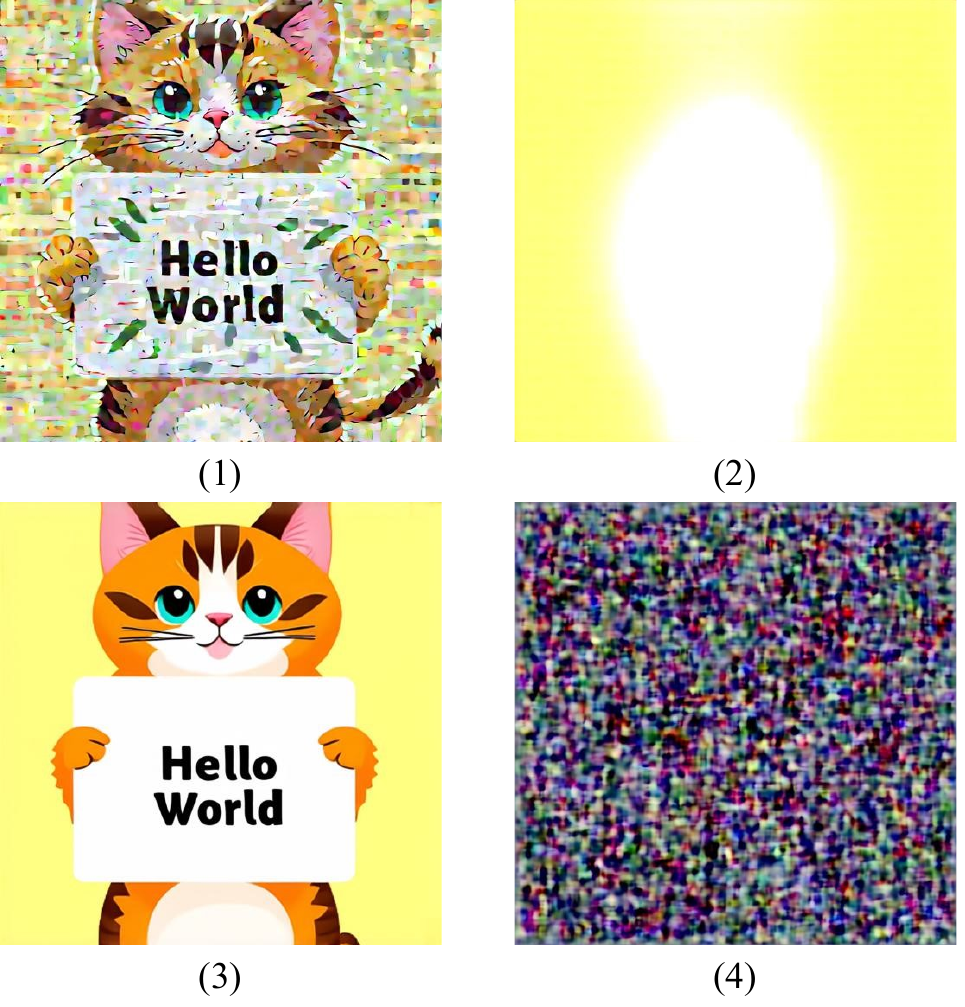}
    \caption{An illustration of the combinations. (1) represents the Combination 1. (2) represents the Combination 2. (3) represents the Combination 3. (4) represents the Combination 4.}
    \label{fig:conbinations}
\end{figure}
We evaluated the influence of these four combinations shown in Fig.~\ref{fig:conbinations}. Combination 1 tended to retain some noise artifacts. This may be because globally modifying $v^{l}_{uc}$ affects noise removal features. Combination 3 focuses on the modification only within the scaled difference term $w*(v^{l*}_{c} - v^{l*}_{uc})$. The result shows that Combination 3 yielded the best results. It effectively reduced oversaturation while better preserving image structure and avoiding the artifacts seen in Combination 1. Therefore, we selected Combination 3 as the optimal for integrating our down-weighting strategy.

Therefore, LF-CFG could be finally formulated as follows:
\begin{equation}
\begin{split}
    dx_{t} &= [ v^{l}_{uc}(t) + \rho w*( m^{l}_{c}(t)v^{l}_{c}(t) - m^{l}_{uc}(t)v^{l}_{uc}(t))\\ 
    &+ w*( (1-m^{l}_{c}(t))v^{l}_{c}(t) - (1-m^{l}_{uc}(t))v^{l}_{uc}(t)) \\
    &+ v^{h}_{uc}(t) + w(v^{h}_{c}(t) - v^{h}_{uc}(t))]dt.
\end{split}
\label{eq:lf-cfg}
\end{equation}

The overall algorithm of LF-CFG is summarized in Algorithm~\ref{al:lf-cfg}. LF-CFG explicitly connects the oversaturation and the CFG, thus removing more redundant information while keeping the valuable part.

\begin{algorithm}[t]
    \caption{ The overall pipeline of LF-CFG} \label{al:lf-cfg}
    \begin{algorithmic}
     \State \textbf{Input:} $v_{\theta}$,$w$,$d(\ast,\ast)$, $f_{l}(\ast)$, $c$, $T$
     \State \textbf{Output:} $x_{0}$
    \State Initialize $x_{T}$. 
    \State $\text{Cache}_{t+1} \gets \text{None}$.
    \State Isfirst $\gets$ True.
    \For{$t$, $t-1$ in pair $[(1,\frac{T-1}{T}),...,(\frac{1}{T},0)]$} 
        \If{Isfirst}
            \State Calculate $v_{uc}(t)$ and $v_{c}(t)$.
            \State $\text{Cache}_{t+1} \gets (v_{uc}(t), v_{c}(t))$.
            \State Calculate $x_{t-1}$ by Eq.~\ref{eq:reverse}.
            \State$ \text{Isfirst} \gets \text{False}$.
            \Else
            \State  Calculate $v_{uc}(t)$ and $v_{c}(t)$.
            \State $\text{Cache}_{t+1} \gets (v_{uc}(t), v_{c}(t))$.
            \State Calculate $x_{t-1}$ by Eq.~\ref{eq:lf-cfg}.
        \EndIf
        \State $x_{t} \gets x_{t-1}$.
    \EndFor
    \State \textbf{Return:} $x_{0}$
\end{algorithmic}
\end{algorithm}

%% file: sec/5_experiment.tex
\section{Experiment}
\label{sec:experiment}
\begin{table*}
    \centering
    \caption{Qualitative comparison among LF-CFG, APG, and CFG based on MS-COCO 2017 under different $w$, where the \textbf{Bold} represents the best results, and the \underline{underline} represents the second best results.}
    \begin{tabular}{ccccccc}
    \toprule
     Model & Guidance & FID $\downarrow$ & KID $\downarrow$ & CLIP $\uparrow$ & Saturation $\downarrow$ \\
     \midrule
    \multirow{3}{*}{ Stable Diffusion 2.1 $(w=5)$} & CFG & \textbf{37.45} & \textbf{0.0082} & 16.75 & 0.32 \\
    & APG &   39.40 &  0.0100 & \underline{16.81} &  \underline{0.31}\\
    & LF-CFG (Ours) & \underline{38.70} & \underline{0.0099} & \textbf{16.83} & \textbf{0.30} \\
         \midrule
    \multirow{3}{*}{ Stable Diffusion 2.1 $(w=10)$} & CFG & 37.36 & 0.0075 & \textbf{16.64} & 0.35 \\
    & APG & \underline{36.17} &  \underline{0.0070} & \underline{16.59} & \underline{0.33}\\
    & LF-CFG (Ours) & \textbf{ 34.00} & \textbf{0.0056} & \textbf{16.64} &\textbf{0.30} \\
         \midrule
    \multirow{3}{*}{ Stable Diffusion 2.1 $(w=15)$} & CFG & 39.85 &  0.0089 & \textbf{16.63} & 0.39 \\
    & APG & \underline{37.45} & \underline{0.0076} & \underline{16.59} & 0.36\\
    & LF-CFG (Ours) & \textbf{33.76} & \textbf{0.0053} & 16.56 & \textbf{ 0.32} \\
         \midrule
    \multirow{3}{*}{ Stable Diffusion-XL $(w=5)$} & CFG & \textbf{91.35} &  \textbf{0.0457}  & \underline{17.125} & \underline{0.31} \\
    & APG &  92.54 & 0.0481 & \textbf{17.31} & \textbf{0.30}\\
    & LF-CFG (Ours) & \underline{91.67} & \underline{0.0465} & \textbf{17.31} & \textbf{0.30} \\
         \midrule
    \multirow{3}{*}{Stable Diffusion-XL $(w=10)$} & CFG & 96.35 &  0.0490 &17.11 &0.34  \\
    & APG & \underline{93.08} &  \underline{0.0464} & \underline{17.12} & \underline{0.32}\\
    & LF-CFG (Ours) & \textbf{ 90.94} & \textbf{0.0445} & \textbf{17.14} & \textbf{ 0.31}\\
         \midrule
    \multirow{3}{*}{Stable Diffusion-XL $(w=15)$} & CFG & 112.68 & 0.0626 & \textbf{17.38} & 0.39 \\
    & APG & \underline{105.83} & 0.0579 & \underline{17.28} & \textbf{0.36}\\
    & LF-CFG (Ours) & \textbf{ 94.20} & \textbf{0.0473} &  17.16 & \underline{0.37} \\
         \midrule
    \multirow{3}{*}{ Stable Diffusion 3 $(w=5)$} & CFG & 37.56 &  0.0088 & 16.48 &0.39 \\
    & APG & \underline{33.71} &  \underline{0.0086} & \underline{16.53} &  \textbf{0.36}\\
    & LF-CFG (Ours) & \textbf{31.21} & \textbf{0.0064} & \textbf{16.78} & \textbf{0.36} \\
        \midrule
    \multirow{3}{*}{ Stable Diffusion 3 $(w=10)$} & CFG & \underline{34.92} & \underline{0.0078} & \textbf{16.67} & 0.44 \\
    & APG &  35.41 &  0.0079 & 16.53 & \underline{0.42}\\
    & LF-CFG (Ours) & \textbf{32.54} & \textbf{0.0075} & \underline{16.64} & \textbf{0.37}\\
         \midrule
    \multirow{3}{*}{ Stable Diffusion 3 $(w=15)$} & CFG & 59.30 & 0.0204 &\textbf{17.55} & 0.48\\
    & APG & \underline{45.02} & \underline{0.0110} & 17.05 &  \underline{0.46}\\
    & LF-CFG (Ours) & \textbf{43.44} & \textbf{ 0.0099} & \underline{17.28} & \textbf{0.45} \\
         \midrule
    \multirow{3}{*}{Stable Diffusion 3.5 $(w=5)$} & CFG & 32.80 &  0.0084 &16.52&0.34  \\
    & APG & \underline{31.46} & \underline{0.0081} & \textbf{16.66} & \underline{0.32}\\
    & LF-CFG (Ours) & \textbf{30.72} & \textbf{0.0074} & 16.58 & \textbf{0.29} \\
         \midrule
    \multirow{3}{*}{  Stable Diffusion 3.5 $(w=10)$} & CFG & 33.58 & \underline{0.0068} & \textbf{16.55}& 0.44 \\
    & APG & \underline{32.67} & 0.0076 & \textbf{16.55} &  \underline{0.41}\\
    & LF-CFG (Ours) & \textbf{30.04} & \textbf{0.0066}& \underline{16.52} & \textbf{ 0.36} \\
         \midrule
    \multirow{3}{*}{ Stable Diffusion 3.5 $(w=15)$} & CFG & 45.35&  0\underline{.0092} & \textbf{17.04} & 0.48 \\
    & APG &  \underline{34.61} & \textbf{0.0061} & 16.75 & \underline{0.47}\\
    & LF-CFG (Ours) & \textbf{33.86} &  \textbf{0.0061} & \underline{16.84} & \textbf{ 0.44} \\
    \bottomrule
    \end{tabular}
    \label{tab:mscoco}
\end{table*}
\textbf{Setup.} We evaluate LF-CFG on the latest diffusion models, including Stable Diffusion 3.0~\cite{sd3}, Stable Diffusion 3.5~\cite{sd3}, and SiT-XL~\cite{sit}. We also evaluate the LF-CFG on Stable Diffusion-XL~\cite{sdxl} and Stable Diffusion 2.1 used in APG to show LF-CFG's improvement further. The main comparison baseline is the APG~\cite{baseline}.

\begin{table*}
    \centering
    \caption{Qualitative comparison among LF-CFG, APG, and CFG based on the ImageNet under different $w$.}
    \begin{tabular}{clccccc}
    \toprule
     Model & Guidance & FID $\downarrow$ & Precision $\uparrow$ & Recall $\uparrow$ & Saturation $\downarrow$ \\
     \midrule
    \multirow{3}{*}{ SiT $(w=3)$} & CFG & 18.29  &  0.75 & 0.71 & 0.36 \\
    & APG & 13.79 &  0.75 & \textbf{0.72} &0.34\\
    & LF-CFG (Ours) & \textbf{6.86} & \textbf{0.76} & \textbf{0.72} & \textbf{0.30} \\
         \midrule
    \multirow{3}{*}{ SiT $(w=7)$} & CFG &  28.14 & 0.76 & 0.54 & 0.43 \\
    & APG & 26.91 &   0.76 & 0.59 &  0.41\\
    & LF-CFG (Ours) & \textbf{ 22.21} & \textbf{0.79} & \textbf{0.68} & \textbf{0.37} \\
         \midrule
    \multirow{3}{*}{ SiT $(w=10)$} & CFG & 28.64 & 0.81 & 0.44 & 0.44 \\
    & APG & 28.84 & 0.83 & 0.52 & 0.42\\
    & LF-CFG (Ours) & \textbf{26.82}& \textbf{0.84} & \textbf{0.60}& \textbf{0.41} \\
    \bottomrule
    \end{tabular}
    \label{tab:imagenet}
\end{table*}

\textbf{Datasets and Evaluation metrics} The pre-trained diffusion models we used were mainly categorized into the label-based condition and text-prompt-based condition. For label-based condition models, we evaluate the performance of LF-CFG on ImageNet~\cite{ImageNet}, where we randomly generate 10K labels from 1K classes to generate 10K images. The metrics include FID~\cite{fid}, precision~\cite{recall_pred}, recall~\cite{recall_pred}, and saturation~\cite{baseline}. Specifically, we choose the validation set of ImageNet as the ground truth to calculate FID, precision, and recall.
\begin{table}
    \centering
    \caption{Ablation study about the super-resolution operator under Stable Diffusion 3 based on MS-COCO 2017 under different $w$, where the \textbf{Bold} represents the best results.}
    \begin{tabular}{ccccccc}
    \toprule
     Scales & FID $\downarrow$ & KID $\downarrow$ & CLIP $\uparrow$ & Saturation $\downarrow$ \\
     \midrule
    $2\times$ & 33.20 & 0.0073 & 16.56 & \textbf{0.37} \\
    $4\times$ &   33.80 & \textbf{0.0071} & \textbf{16.73} & 0.40\\
    $8\times$ & \textbf{32.54} & 0.0075 & 16.64 & \textbf{0.37} \\
    \bottomrule
    \end{tabular}
    \label{tab:ala_resolution}
\end{table}

For text prompt-based condition models, we evaluate the performance of LF-CFG on MS COCO 2017~\cite{mscoco}, where we randomly choose 10K from the captions of MS COCO 2017 to generate 10K images. The metrics includes FID~\cite{fid}, KID~\cite{kid}, CLIP~\cite{clip}, and saturation~\cite{baseline}. The ground truth for calculating FID is the 10K images from the MS COCO dataset related to the caption we chose. Then, we use KID to replace precision and recall since they are sensitive to the hyperparameter setting, such as the cluster number. Additionally, we add the CLIP metric, i.e., the clip score between the text prompt and generated images, to measure its generation quality.

\textbf{Hyperparameter settings.} To make a fair comparison, we set $T=20$, the time step for all models. Then, we keep the default setting for the rest of the hyperparameters involved in the baselines and pre-trained diffusion models. Specifically, Stable Diffusion-XL~\cite{sdxl} and Stable Diffusion 2.1 belong to the diffusion probabilistic models (DPM) based paradigm~\cite{ddpm}. The rest belong to the flow-match-based models. 

\textbf{Quantitative results.} We report the qualitative results shown in Table~\ref{tab:imagenet} and Table~\ref{tab:mscoco}. These results first prove that the LF-CFG could alleviate the oversaturation. It could be found that LF-CFG further improves the saturation metric compared to APG. Such improvement illustrates that the redundant information hidden on the low-frequency signal will be accumulated during the generation process, which finally leads to oversaturation.

Alleviating the oversaturation ensures LF-CFG could generate high-quality images under a high $w$. Concretely, LF-CFG achieves the best average of 18.66, 0.41, and 0.82 FID, precision, and recall for label-based condition generation, respectively. Similar results could be obtained for text prompts-based condition generation, except for the Stable Diffusion ($w=5$), LF-CFG performs best on the FID and KID metrics. 
\begin{figure*}
    \centering
    \includegraphics[width=0.9\linewidth]{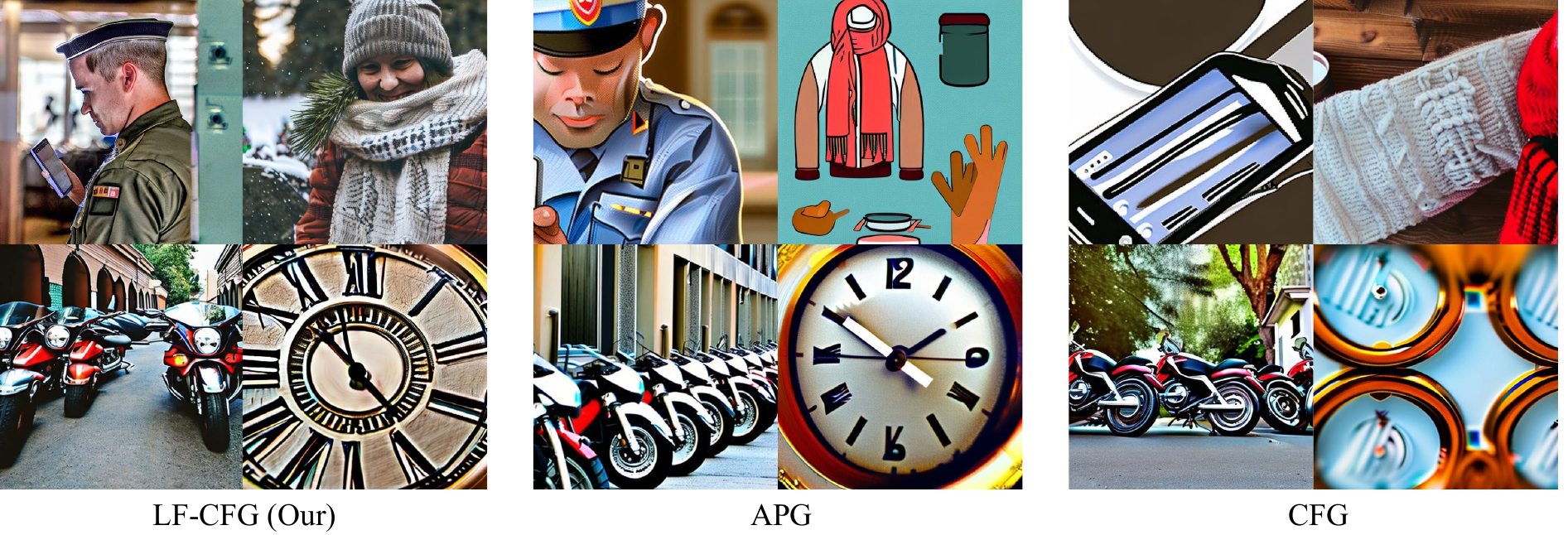}
    \caption{Qualitative results based on Stable Diffusion 2.1 ($w=15$)}
    \label{fig:sd2.1}
\end{figure*}
LF-CFG further shows its potential to maintain the text-image alignment on the text prompt-based condition generation. Our experimental results illustrate that LF-CFG could keep the loss of the text-image alignment under a suitable level. Concretely, except in two situations, LF-CFG achieves the best performance compared to the APG and the second best performance compared to the CFG. Meanwhile, compared to the decrease in CLIP score, LF-CFG significantly improved image generation quality, thus leading to a better trade-off than the APG. 

\textbf{Qualitative results.} To further illustrate the superiority of LF-CFG, we visualize the generated images among CFG, APG, and LF-CFG under different diffusion models, and the highest $w$ shown in Fig.~\ref{fig:sd3} - \ref{fig:sd3.5}. The supplementary shows more results under different $w$ with Stable Diffusion 2.1 and XL. Specifically, LF-CFG alleviates the oversaturation to improve the generation quality under high $w$, which proves the validity of LF-CFG.

\begin{table}
    \centering
    \caption{The computation cost under Stable Diffusion 3 via RTX 4090 for generating a single image, where MEM is the memory cost.}
    \begin{tabular}{ccc}
    \toprule
     \textbf{Method}& \textbf{Times} (s) $\downarrow$ & \textbf{MEM} (GB) $\downarrow$ \\
     \midrule
    CFG & \textbf{1.03} & \textbf{19.5}\\
    APG &   1.06 &  19.6\\
   LF-CFG & 1.05 & 19.6 \\
    \bottomrule
    \end{tabular}
    \label{tab:time_cost}
\end{table}

\textbf{Computation cost.} To prove the efficiency of the LF-CFG, we report the computation cost of LF-CFG shown in Table~\ref{tab:time_cost}. The results show that LF-CFG will not dramatically increase the computation cost.
\begin{table*}
    \centering
    \caption{Ablation study about $\gamma$ under Stable Diffusion 3 based on MS-COCO 2017 with different $w$, where the \textbf{Bold} represents the best results.}
    \begin{tabular}{ccccccc}
    \toprule
     Method& FID $\downarrow$ & KID $\downarrow$ & CLIP $\uparrow$ & Saturation $\downarrow$ \\
     \midrule
     $\text{mean}(r^{l}_{j}(t)) + \text{std}(r^{l}_{j}(t))$($w=5$) & \textbf{31.21} & \textbf{0.0064} & \textbf{16.78} & \textbf{0.36} \\
    $\text{mean}(r^{l}_{j}(t)) + 2\text{std}(r^{l}_{j}(t))$ ($w=5$)&   31.50 &  0.0066 &  16.64 & 0.37\\
    $\text{mean}(r^{l}_{j}(t)) + 3\text{std}(r^{l}_{j}(t))$($w=5$)& 31.62 &  0.0068 & 16.69 & 0.37 \\
    \midrule
   $\text{mean}(r^{l}_{j}(t)) + \text{std}(r^{l}_{j}(t))$ ($w=10$) &\textbf{32.54} & 0.0075 & \textbf{16.64} & \textbf{0.37} \\
    $\text{mean}(r^{l}_{j}(t)) + 2\text{std}(r^{l}_{j}(t))$ ($w=10$) &  33.07 & \textbf{0.0074} & 16.47 & \textbf{0.37}\\
    $\text{mean}(r^{l}_{j}(t)) + 3\text{std}(r^{l}_{j}(t))$ ($w=10$) &  33.03 & \textbf{0.0074} &  16.50 & \textbf{0.37} \\
    $\text{mean}(r^{l}_{j}(t)) - \text{std}(r^{l}_{j}(t))$ ($w=10$)& 37.73 &0.0078&  16.59 &  0.44\\
    \bottomrule
    \end{tabular}
    \label{tab:ala_redudant}
\end{table*}

\textbf{Comparison to the enhanced method.} To show that the LF-CFG could be compatible to other enhanced methods. We report the additional experiments based on PAG~\ref{tab:pag} shown in Table~\ref{tab:pag}. It can be found that PAG with LF-CFG outperforms the PAG in the optimal $w$. This demonstrates the necessity for further exploring the oversaturation and the validity of LF-CFG.
\begin{table}
    \centering
    \caption{Comparison to the enhanced method under Stable Diffusion-XL based on MS-COCO 2017 with different $w$, where the \textbf{Bold} represents the best results and $\ast$ represents the optimal setting in the original paper.}
    \begin{tabular}{ccccccc}
    \toprule
     \textbf{Method} & \textbf{FID} $\downarrow$ & \textbf{KID} $\downarrow$ & \textbf{CLIP} $\uparrow$ & \textbf{Saturation} $\downarrow$ \\
     \midrule
    PAG ($w=5$)$\ast$ & 90.98  & 0.0412 & 17.23 & \textbf{0.33} \\
    PAG ($w=10$) &   111.13 & 0.0604 & 17.39 & 0.36\\
    PAG ($w=15$) &121.71 &  0.0693 & 17.61&0.39\\
    PAG ($w=10$) + LF-CFG& \textbf{88.52} & \textbf{0.0361} & \textbf{17.62} & \textbf{0.33}\\
    \bottomrule
    \end{tabular}
    \label{tab:pag}
\end{table}

\subsection{Ablation Study}
\textbf{Scale of super-resolution operator.} To explore the influence of different scales of super-resolution operators for extracting low-frequency signals, we report an ablation study shown in Table~\ref{tab:ala_resolution}. The results demonstrate that LF-CFG could work for different scales, and we chose the $8\times$ based on our experiments.

\textbf{The ratio of the redundant information.} Table~\ref{tab:ala_redudant} reports the influence of the redundant information ratio $\rho$. $\rho$ is decided by the statistical principle based on the $\text{mean}(r^{l}_{j}(t))$ and $\text{std}(r^{l}_{j}(t))$ since they decide the ratio between high- and low- change regions. Concretely, $\text{mean}(r^{l}_{j}(t)) + \text{std}(r^{l}_{j}(t))$ indicates $\rho=33.3\%/66.7\%$, where there are $66.7\%$ values fall into the low-change regions. $\text{mean}(r^{l}_{j}(t)) + 2\text{std}(r^{l}_{j}(t))$ indicates $\rho \approx 10.0\%/90.\%$. Additional, we also explore $\text{mean}(r^{l}_{j}(t)) - \text{std}(r^{l}_{j}(t))$ i.e., our initial roughly spiting way. The results show that $\text{mean}(r^{l}_{j}(t)) + \text{std}(r^{l}_{j}(t))$ achieves the best results.

\textbf{Comparison to the unconditional generation.} We report the additional experiments for unconditional generation shown in the supplementary. It could be found: 1) CFG could improve the generation quality, proving the validity of our experiments. 2) LF-CFG could further improve CFG.

\textbf{Comparison to the different solvers.} To prove that LF-CFG could work on different high-order solvers such as DPM-Solver~\cite{dpm}. We report the additional experiment in the supplementary. The results demonstrate that LF-CFG works on different solvers.

\textbf{Comparison on the pixel space.} Stable Diffusion based on the latent space. To verify that LF-CFG could work on pixel space too, we report the experiments based on EDM2~\cite{edm2}, i.e., the pixel space diffusion model, reported in the supplementary. The results show that LF-CFG could improve CFG on the pixel space.

%% file: sec/6_conclusion.tex
\section{Conclusion}
\label{sec:con}
We introduce LF-CFG, the low-frequency improved classifier-free guidance. LF-CFG is based on a new insight into low-frequency signals: redundant information is hidden within these signals and accumulates, causing pixel values to become excessively high. To address this issue, LF-CFG proposes a novel method to identify such redundant information and then apply a down-weighting strategy to weaken these signals without affecting others. Experimental results show that LF-CFG successfully alleviates oversaturation and improves generation quality.

\textbf{Limitation.} The low-frequency signal influences image pixels and other information, such as denoising Gaussian noise. Consequently, LF-CFG may inadvertently weaken both redundant and beneficial information. This aspect requires further investigation in future studies.

%% file: sec/x_supply.tex
\newpage
\appendix
\section{Appendix}
\begin{figure*}
    \centering
    \includegraphics[width=0.9\linewidth]{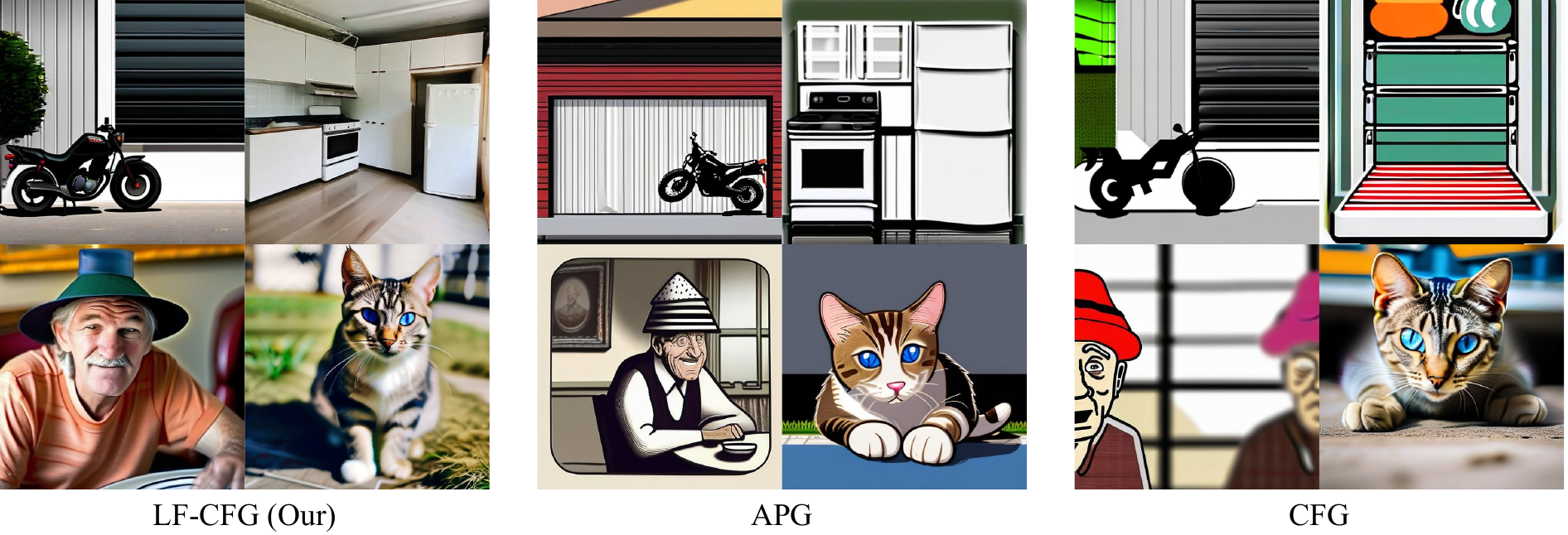}
    \caption{Qualitative results based on Stable Diffusion-XL ($w=15$)}
    \label{fig:sdxl}
\end{figure*}
\subsection{Ablation Study}
This section first shows that the LF-CFG could be compatible with various enhancement methods such as CADS. Then, we add additional results about the unconditional generation to verity the validity of our experiments. We also report additional experiments to show that LF-CFG could work on high-order solvers. In the end, we report additional experiments to prove that LF-CFG could work on pixel space, too.

\textbf{Compatible with enhanced methods.} To illustrate that LF-CFG is compatible with the noise scheduler method, such as CADS, we report the ablation study in Table~\ref{tab:ala_cads}, where the CADS could further improve LF-CFG.

\begin{table*}
    \centering
    \caption{Ablation study of compatible under Stable Diffusion 3 based on MS-COCO 2017.}
    \begin{tabular}{ccccccc}
    \toprule
     Method& FID $\downarrow$ & KID $\downarrow$ & CLIP $\uparrow$ & Saturation $\downarrow$ \\
     \midrule
     Stable Diffusion ($w=10$) & 32.54 & 0.0075 & 16.64 & 0.37 \\
     Stable Diffusion ($w=10$) + CADS &  30.76 &  0.0045 & 16.95 & 0.37 \\
    \bottomrule
    \end{tabular}
    \label{tab:ala_cads}
\end{table*}

\textbf{Comparison to the unconditional generation.} To further show how LF-CFG improves CFG, we report the additional unconditioned generation results shown in Table~\ref{tab:ala_unc}. The results ensure that CFG could improve the generation quality while LF-CFG further improves the CFG.

\begin{table*}
    \centering
    \caption{Ablation study of unconditional generation on MS-COCO 2017. Since $w=0$ is unconditioned generation, there is no CLIP score. }
    \begin{tabular}{ccccccc}
    \toprule
     Method& FID $\downarrow$ & KID $\downarrow$ & CLIP $\uparrow$ & Saturation $\downarrow$ \\
     \midrule
    Stable Diffusion 2.1 ($w=0$) & 124.56  &  0.0962 & - & 0.37 \\
     Stable Diffusion-XL ($w=0$) &   162.64 &  0.1150 &-  & 0.37 \\
     Stable Diffusion 3.0 ($w=0$) &   43.67 & 0.0174 & - &0.36 \\
    Stable Diffusion 3.5 ($w=0$) &  38.06  &  0.0092 & -  & 0.26 \\
    \bottomrule
    \end{tabular}
    \label{tab:ala_unc}
\end{table*}

\textbf{Ablation study of different solvers.} To illustrate that LF-CFG could work on different solvers, such as high-order solvers, we report the experiments shown in Table~\ref{tab:ala_solver}. The results show that LF-CFG works well on DPM-Solver.
\begin{table*}
    \centering
    \caption{Ablation study of different solvers with Stable Diffusion 2.1 on MS-COCO 2017, where $\ast$ means the setting in the main paper.}
    \begin{tabular}{ccccccc}
    \toprule
     Method& FID $\downarrow$ & KID $\downarrow$ & CLIP $\uparrow$ & Saturation $\downarrow$ \\
     \midrule
    DDIM ($w=10$)$\ast$ & 34.00 & 0.0056 & 16.64 &  0.30  \\
     DPM-Solver ($w=10$) &  34.51 & 0.0063&16.82& 0.31 |  \\
    \bottomrule
    \end{tabular}
    \label{tab:ala_solver}
\end{table*}

\textbf{Comparison on the pixel space.} We add the comparisons based on EDM2 (pixel-level based diffusion model). The checkpoint is EDM2-S (ImageNet 512 $\times$ 512). We keep the same hyperparameter setting in APG to make a fair comparison. The results are shown in Table~\ref{tab:ala_edm}. The results prove that LF-CFG could further improve CFG compared to the APG, which verifies that LF-CFG could work on pixel space.

\begin{table*}
    \centering
    \caption{Ablation study of pixel space with EDM2 on ImageNet, where $\ast$ means the setting in the main paper.}
    \begin{tabular}{ccccccc}
    \toprule
     Method& FID $\downarrow$ & KID $\downarrow$ & CLIP $\uparrow$ & Saturation $\downarrow$ \\
     \midrule
        CFG ($w=4$) &26.35 &0.63& 0.32& 0.54 \\ 
        APG ($w=4$) &23.32 & 0.67&0.48& 0.43 \\  
        LF-CFG ($w=4$) &\textbf{22.28}& \textbf{0.71}& \textbf{0.67}& \textbf{0.40}\\ 
    \bottomrule
    \end{tabular}
    \label{tab:ala_edm}
\end{table*}

\subsection{More Results}
\begin{figure*}
    \centering
    \begin{subfigure}[b]{0.45\textwidth}
    \includegraphics[width=\textwidth,scale=1.1]{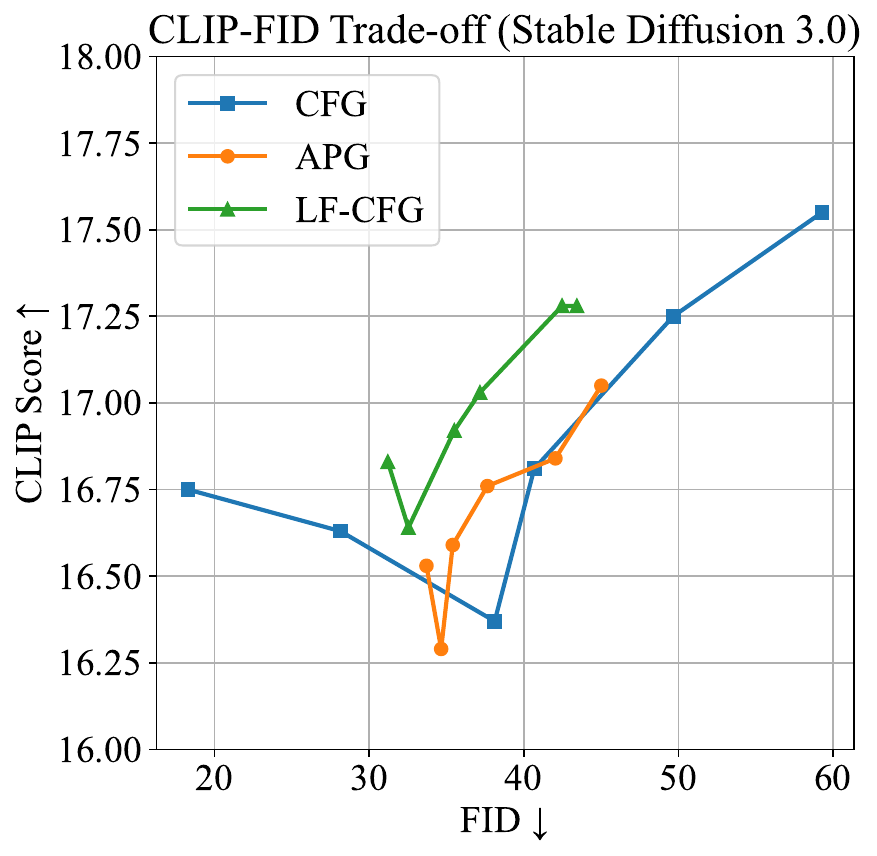}
    \label{fig:1}
    \caption{CLIP-FID curve via Stable Diffusion 3.0. The $w = \{5,7,9,11,13,15\}$.}
    \end{subfigure}
    \begin{subfigure}[b]{0.45\textwidth}
    \includegraphics[width=\textwidth,scale=1.1]{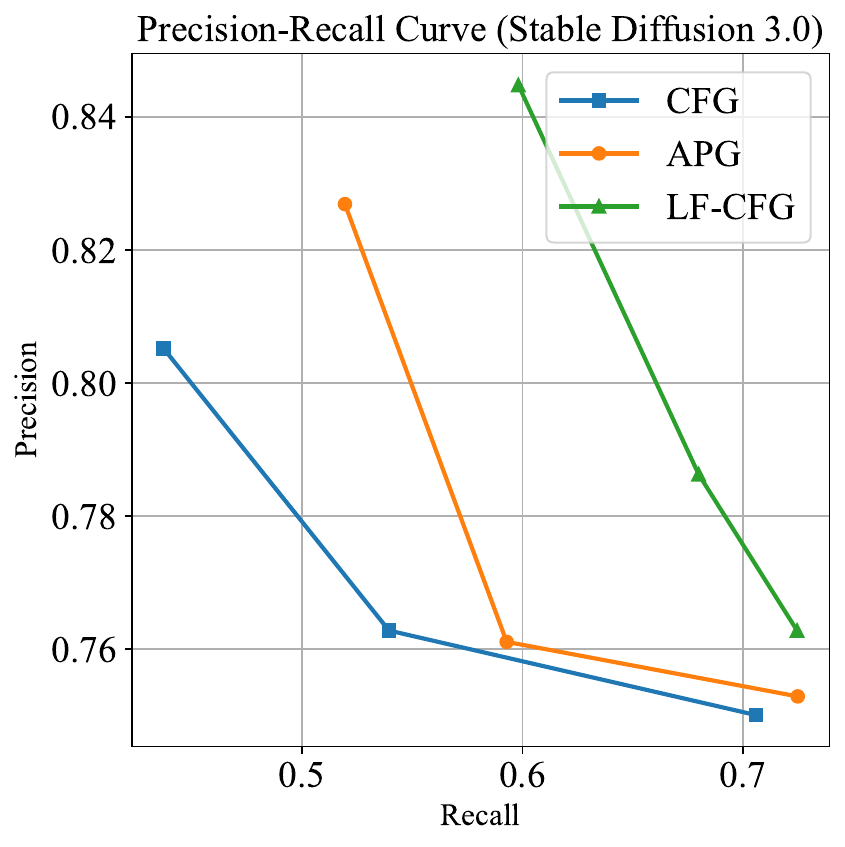}
    \label{fig:1}
    \caption{P-R curve via SiT.The $w = \{3,7,10\}$.}
    \end{subfigure}
    \caption{ The additional experiments for P-R curve and CLIP-FID curve. It could be found that LF-CFG achieves the best text-image alignment. Meanwhile, LF-CFG also achieves the best trade-off between diversity and quality.}
    \label{fig:hunyuan_layer_attention_score_eav}
\end{figure*}

\begin{figure*}
    \centering
    \includegraphics[width=0.9\linewidth]{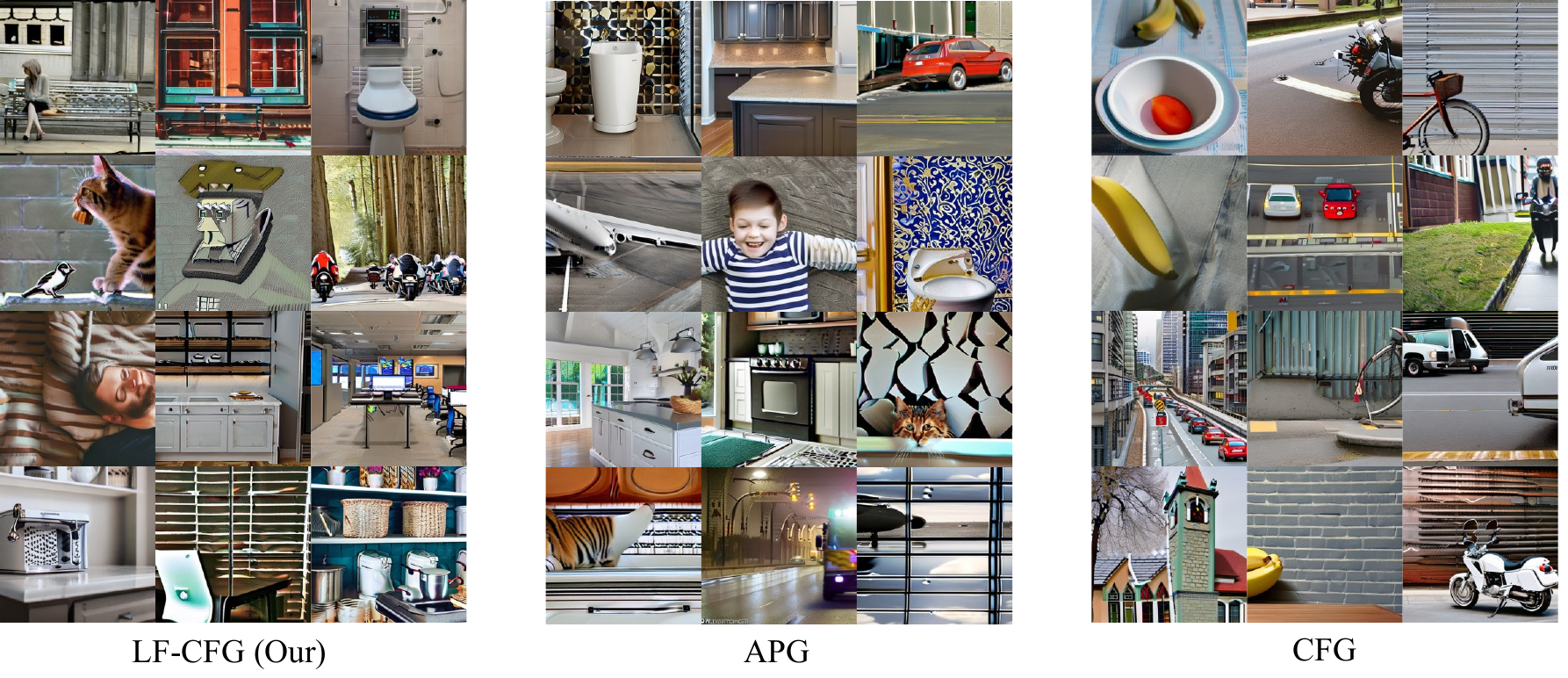}
    \caption{Extra qualitative results based on Stable Diffusion 2.1 ($w=5$)}
    \label{fig:sdxl}
\end{figure*}

\begin{figure*}
    \centering
    \includegraphics[width=0.9\linewidth]{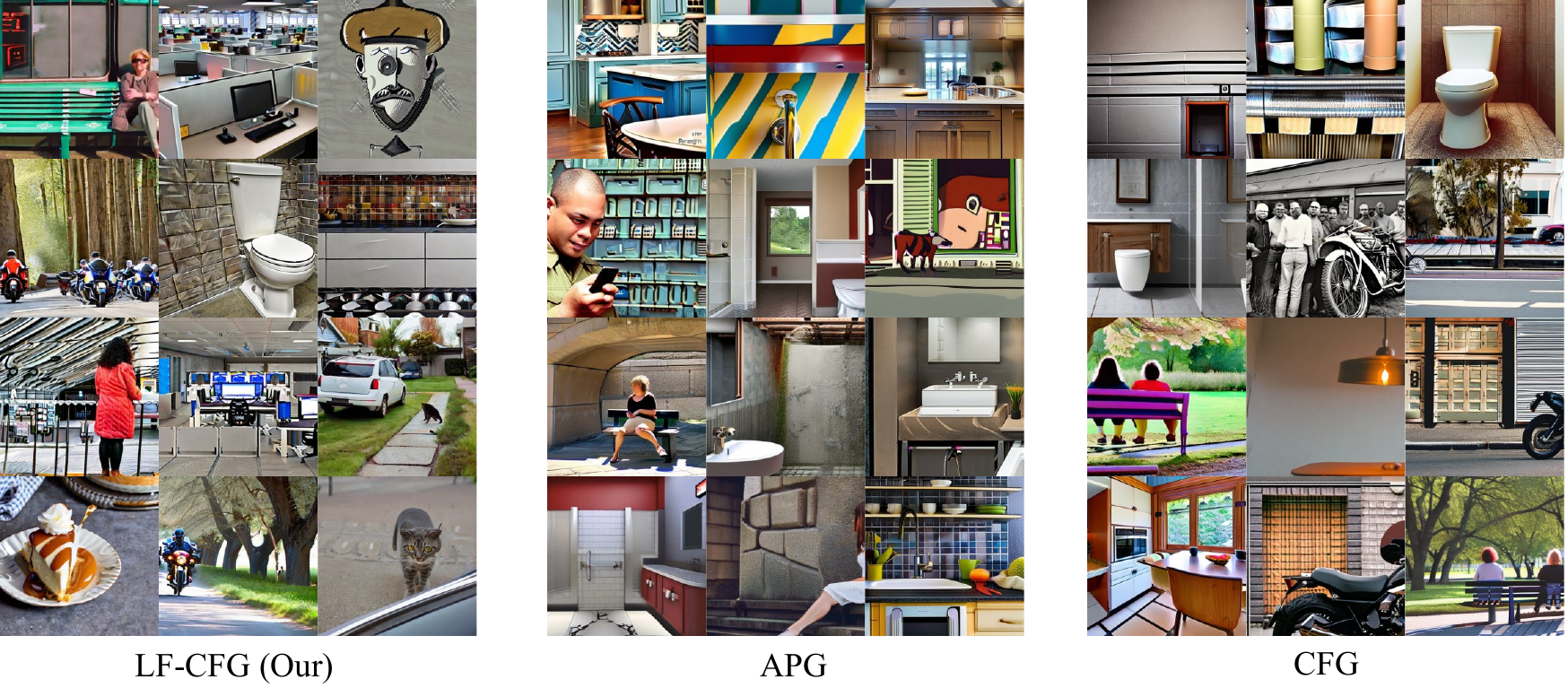}
    \caption{Extra qualitative results based on Stable Diffusion 2.1 ($w=10$)}
    \label{fig:sdxl}
\end{figure*}

\begin{figure*}
    \centering
    \includegraphics[width=0.9\linewidth]{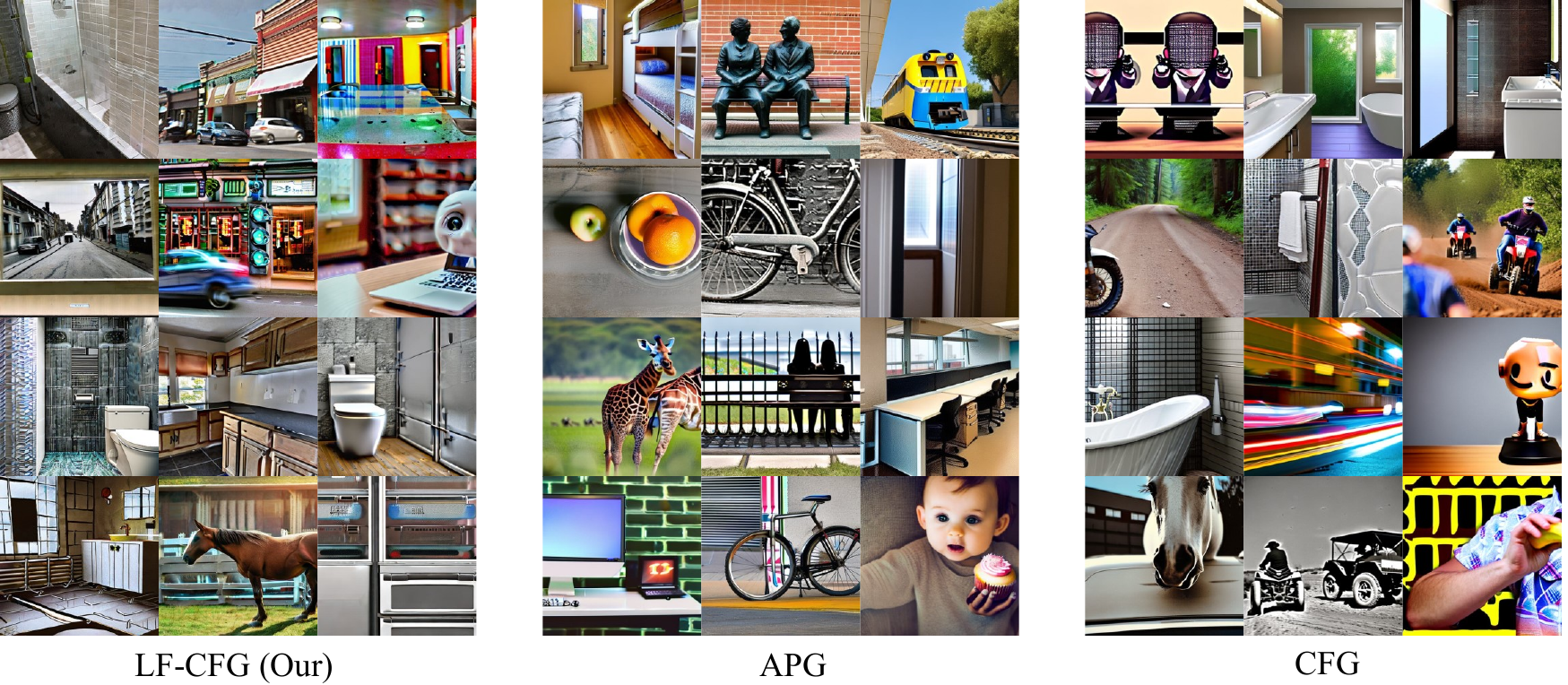}
    \caption{Extra qualitative results based on Stable Diffusion 2.1 ($w=15$)}
    \label{fig:sdxl}
\end{figure*}

\begin{figure*}
    \centering
    \includegraphics[width=0.9\linewidth]{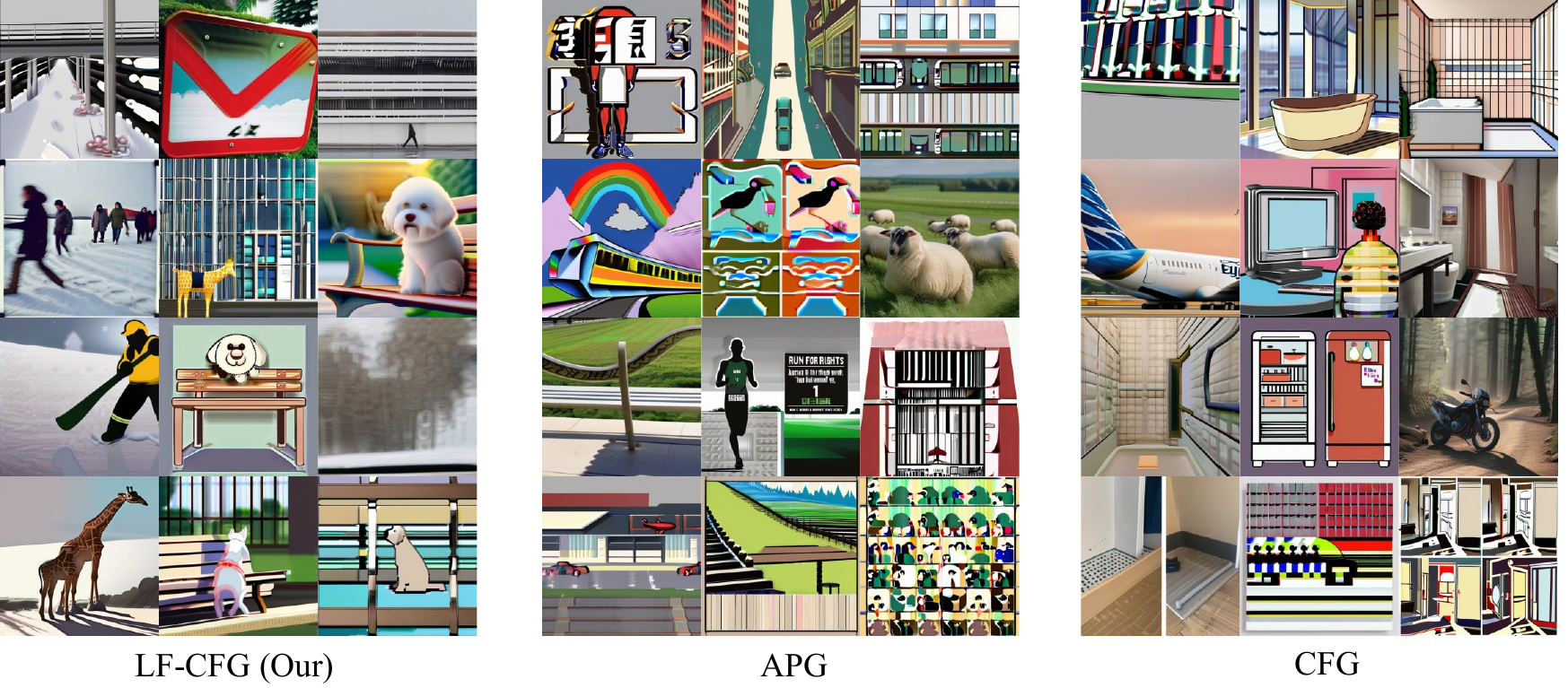}
    \caption{Extra qualitative results based on Stable Diffusion-XL ($w=5$)}
    \label{fig:sdxl}
\end{figure*}

\begin{figure*}
    \centering
    \includegraphics[width=0.9\linewidth]{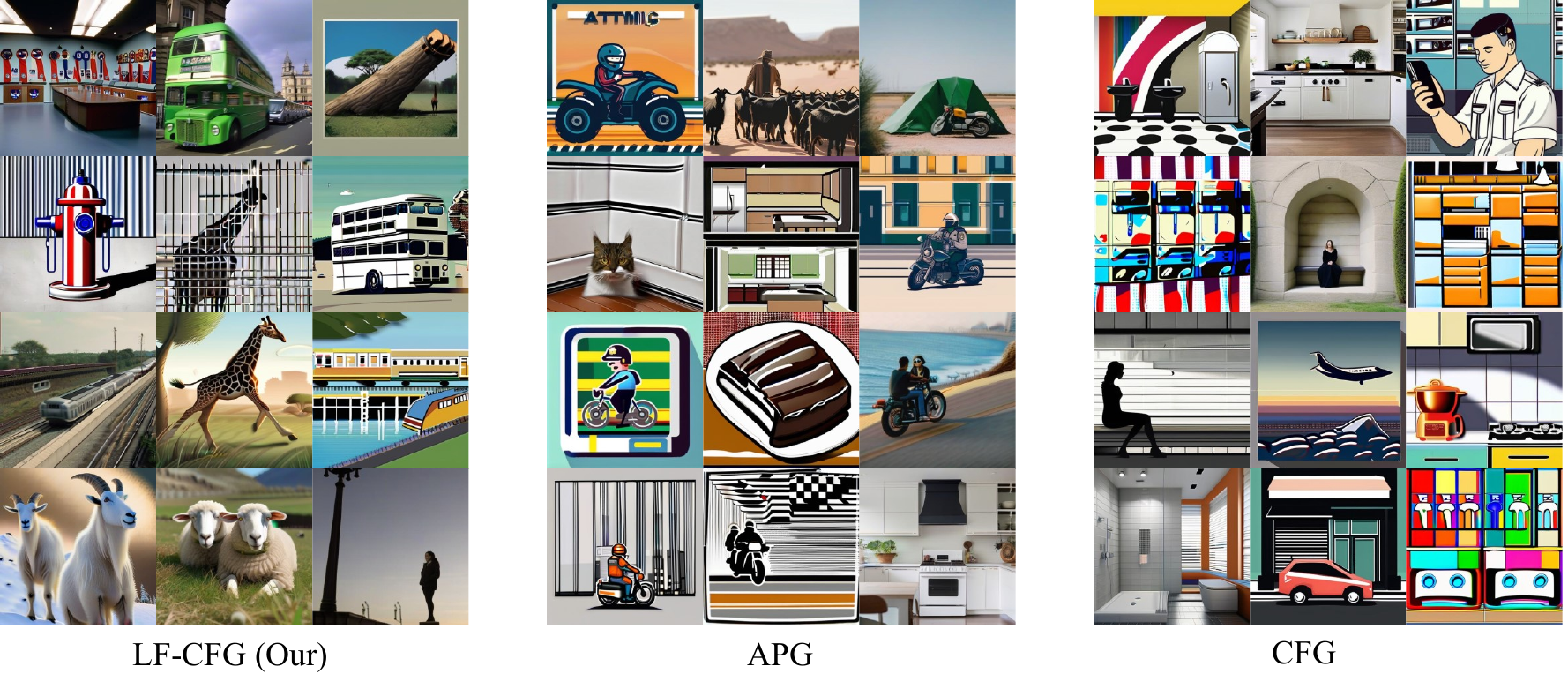}
    \caption{Extra qualitative results based on Stable Diffusion-XL ($w=10$)}
    \label{fig:sdxl}
\end{figure*}

\begin{figure*}
    \centering
    \includegraphics[width=0.9\linewidth]{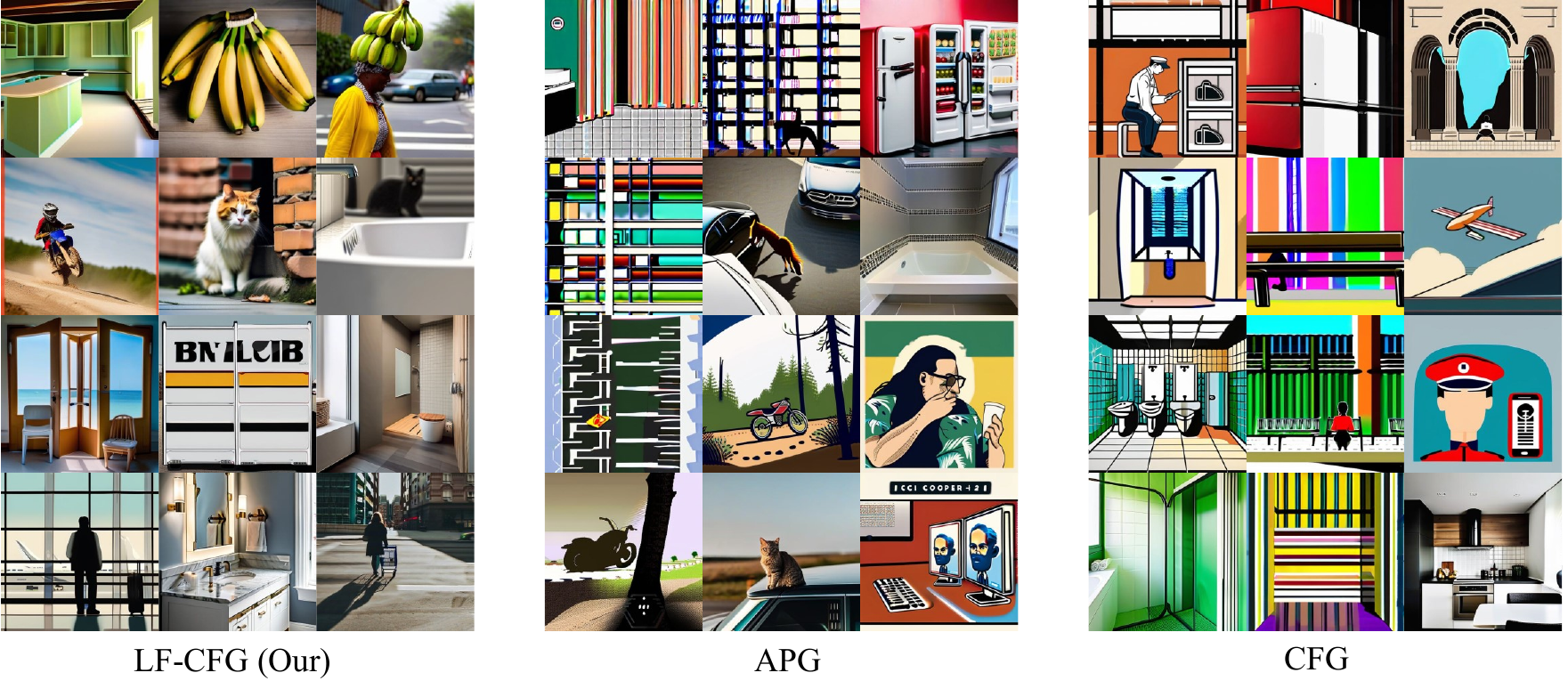}
    \caption{Extra qualitative results based on Stable Diffusion-XL ($w=15$)}
    \label{fig:sdxl}
\end{figure*}

\begin{figure*}
    \centering
    \includegraphics[width=0.9\linewidth]{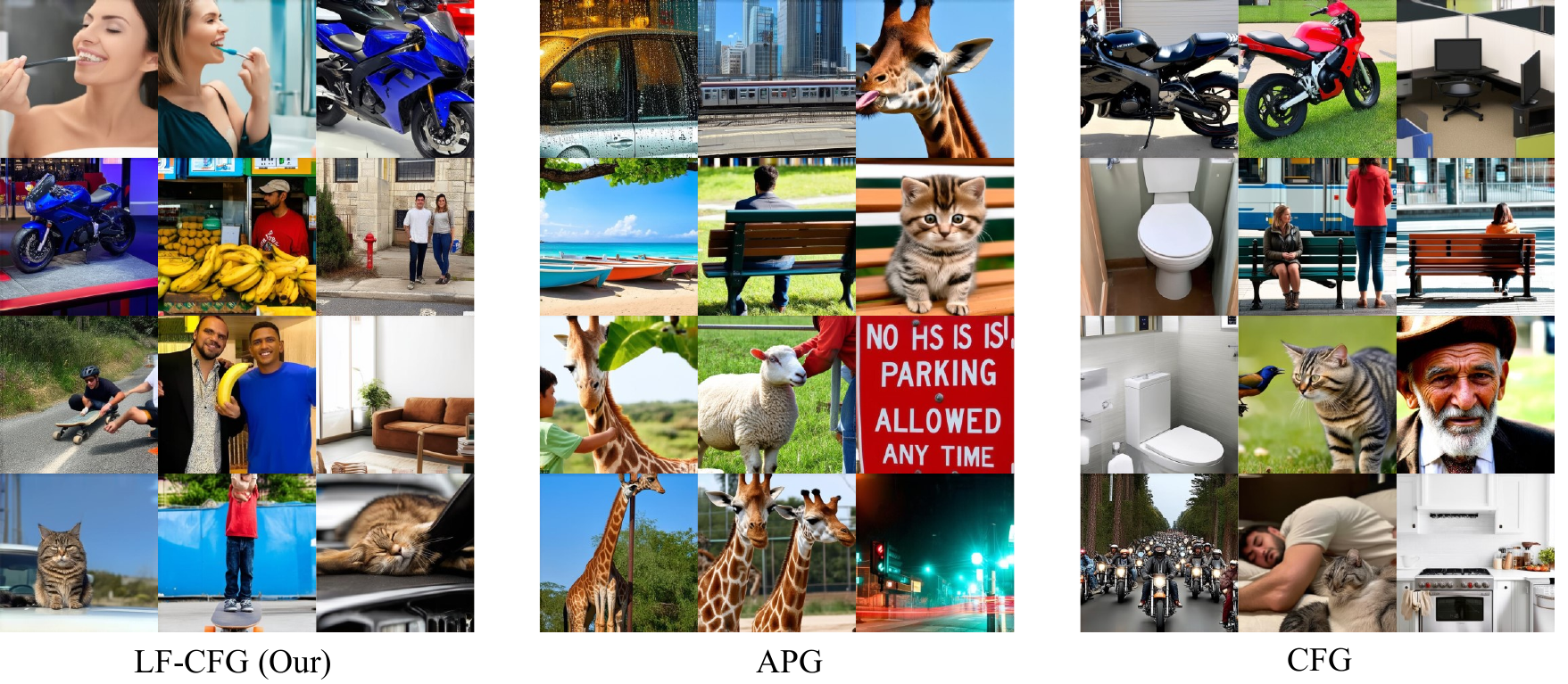}
    \caption{Extra qualitative results based on Stable Diffusion 3 ($w=5$)}
    \label{fig:sdxl}
\end{figure*}

\begin{figure*}
    \centering
    \includegraphics[width=0.9\linewidth]{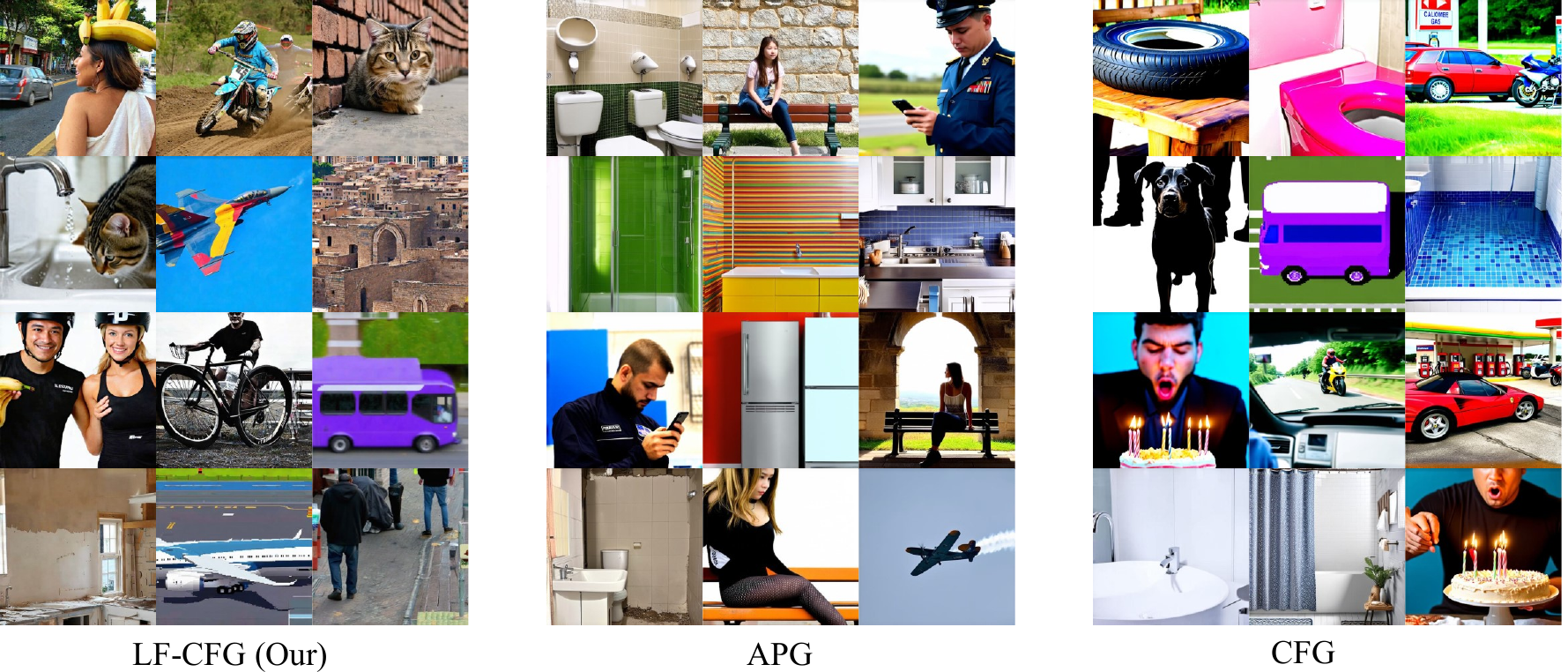}
    \caption{Extra qualitative results based on Stable Diffusion 3 ($w=10$)}
    \label{fig:sdxl}
\end{figure*}

\begin{figure*}
    \centering
    \includegraphics[width=0.9\linewidth]{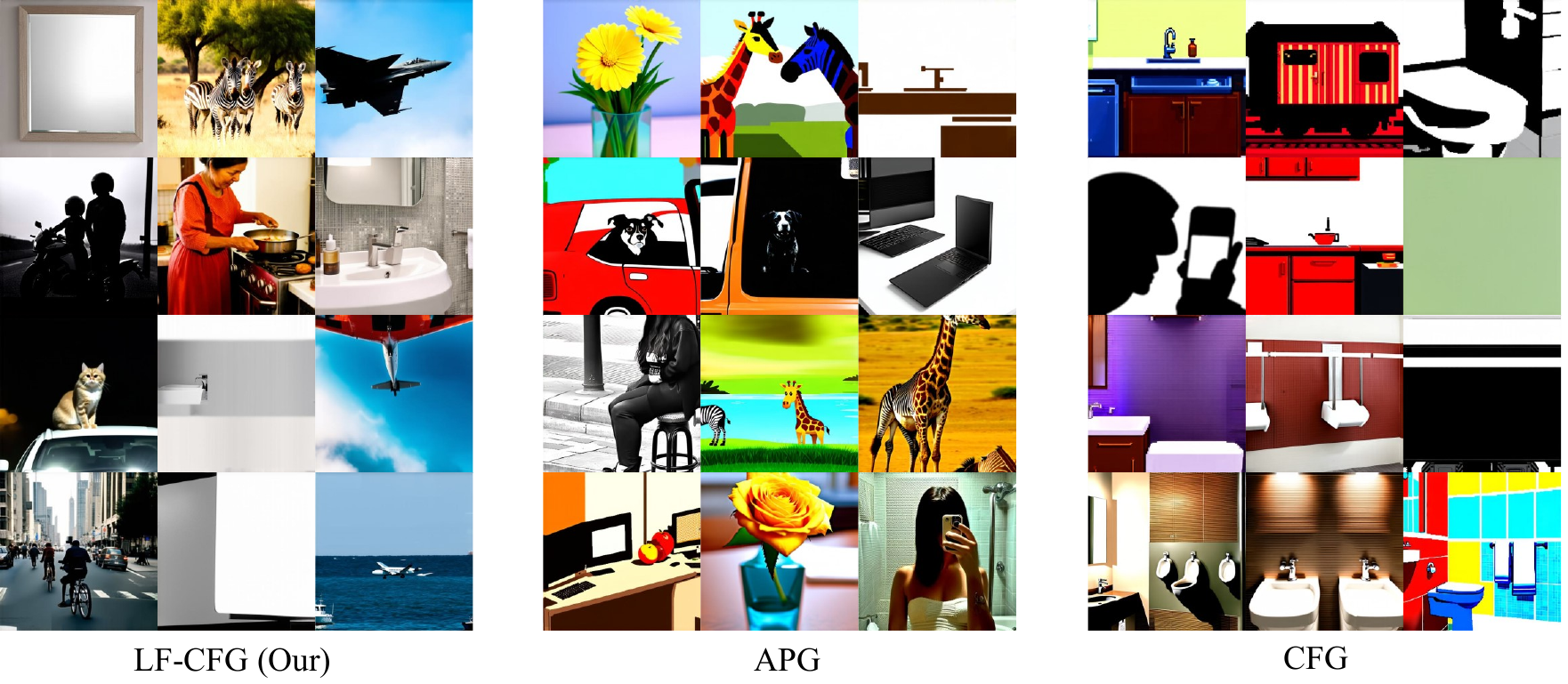}
    \caption{Extra qualitative results based on Stable Diffusion 3 ($w=15$)}
    \label{fig:sdxl}
\end{figure*}

\begin{figure*}
    \centering
    \includegraphics[width=0.9\linewidth]{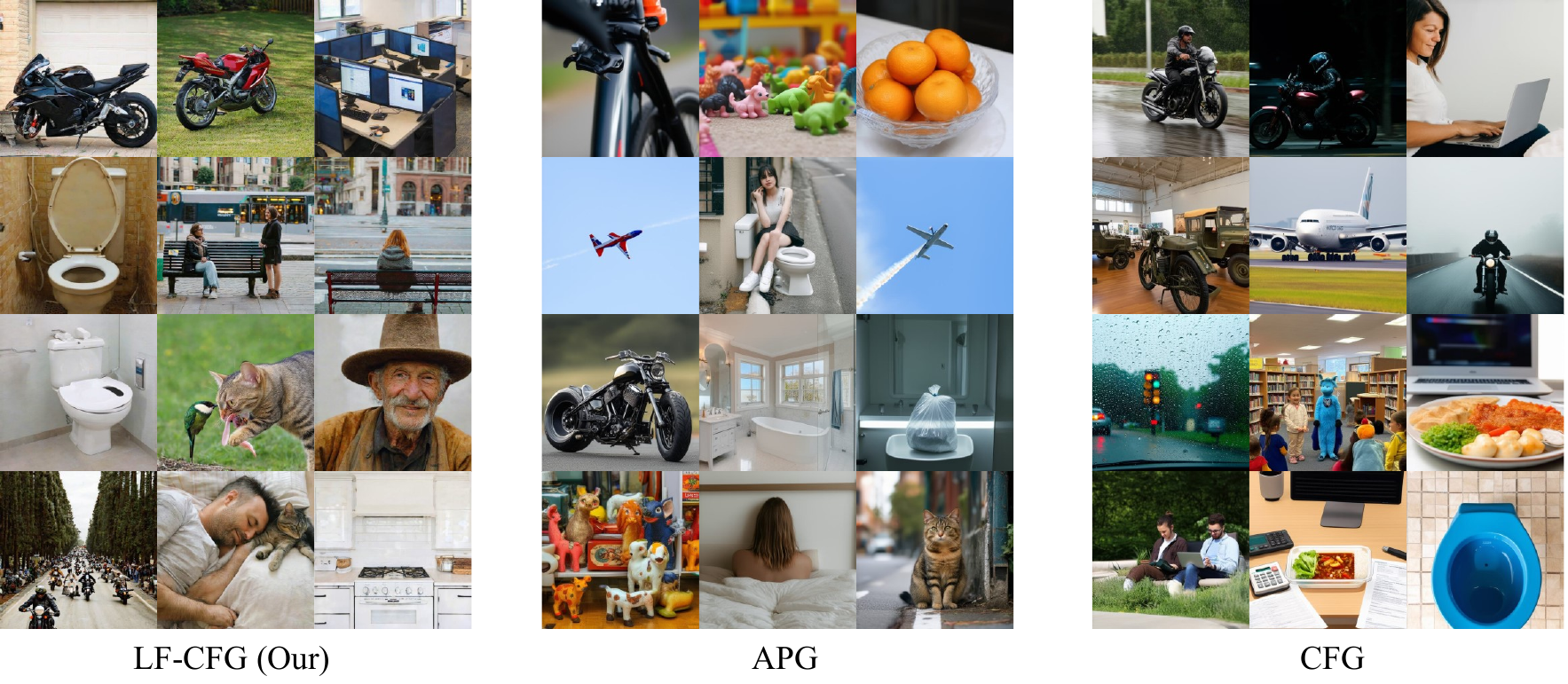}
    \caption{Extra qualitative results based on Stable Diffusion 3.5 ($w=5$)}
    \label{fig:sdxl}
\end{figure*}

\begin{figure*}
    \centering
    \includegraphics[width=0.9\linewidth]{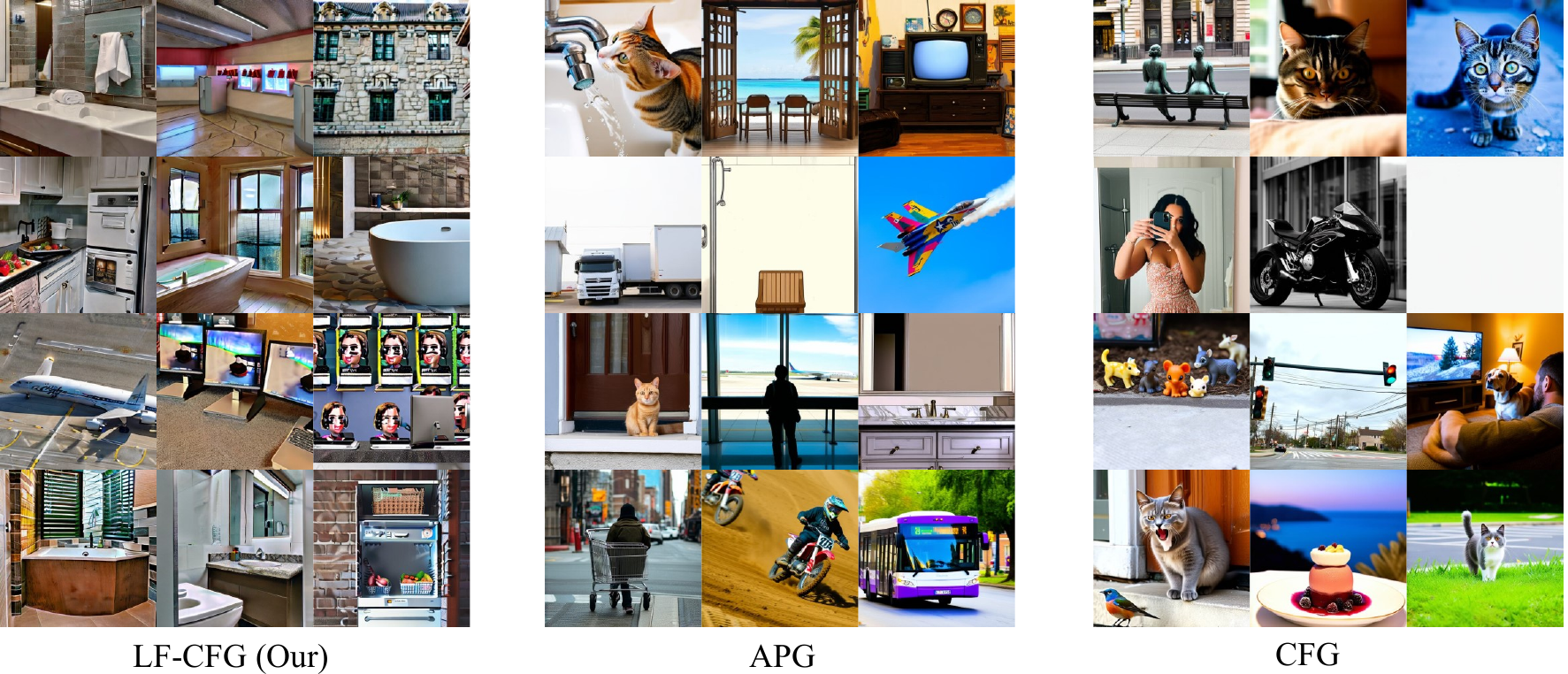}
    \caption{Extra qualitative results based on Stable Diffusion 3.5 ($w=10$)}
    \label{fig:sdxl}
\end{figure*}

\begin{figure*}
    \centering
    \includegraphics[width=0.9\linewidth]{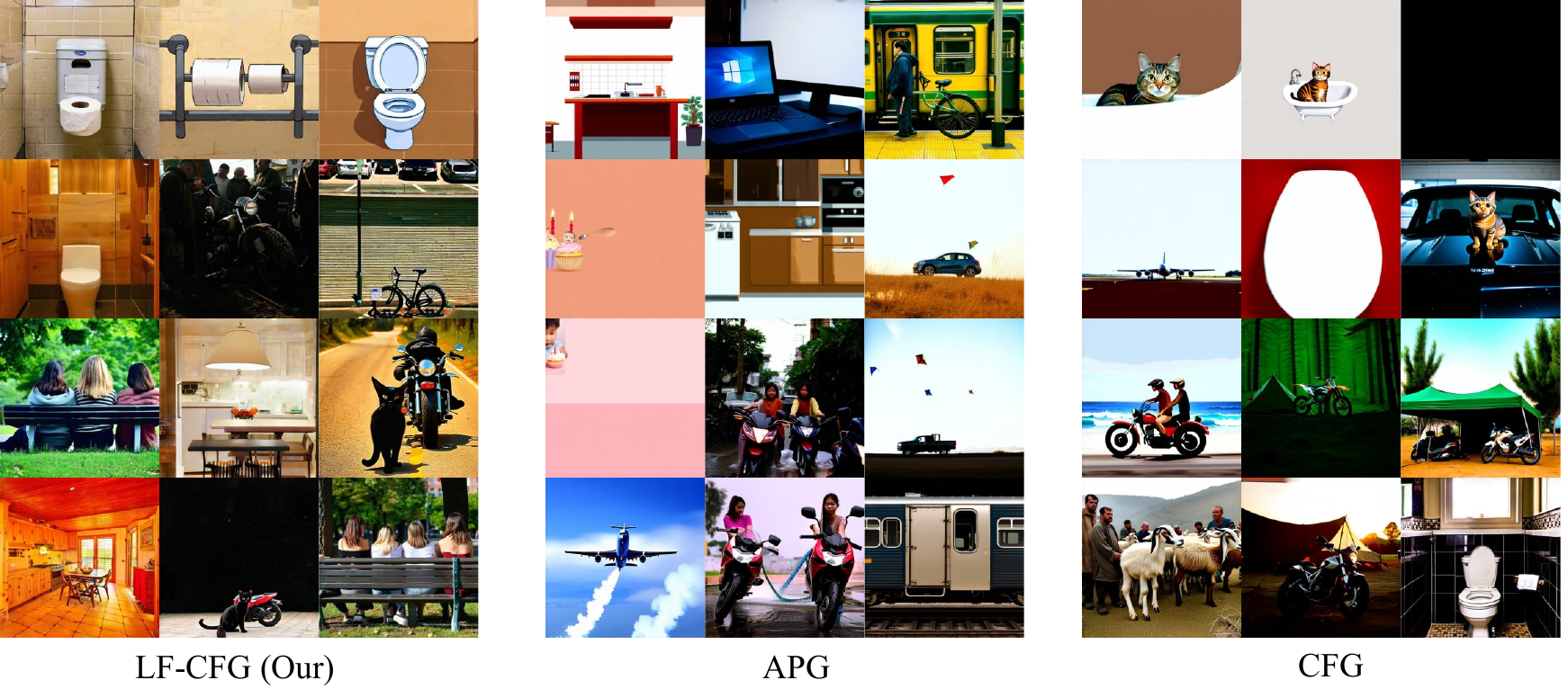}
    \caption{Extra qualitative results based on Stable Diffusion 3.5 ($w=15$)}
    \label{fig:sdxl}
\end{figure*}